\documentclass[conference]{IEEEtran}
\ifCLASSINFOpdf
\else
\fi
\usepackage{graphicx}

\usepackage{amsmath}
\usepackage{caption}
\usepackage{subcaption}
\usepackage{url}
\usepackage{hyperref} 
\usepackage{gensymb}

\usepackage{amssymb}
\usepackage[T1]{fontenc}
\usepackage[utf8]{inputenc}

\usepackage{placeins}
\usepackage{float}
\restylefloat{table}

\begin{document}

\title{\textbf{Cooper: Cooperative Perception for Connected Autonomous Vehicles based on 3D Point Clouds}}

\author{
  \IEEEauthorblockN{
    Qi Chen\IEEEauthorrefmark{1},
    Sihai Tang\IEEEauthorrefmark{1},
    Qing Yang\IEEEauthorrefmark{2} and
    Song Fu\IEEEauthorrefmark{2}
  }
  \IEEEauthorblockA{
    \textit{Department of Computer Science and Engineering} \\
    \textit{University of North Texas, USA}
  }
  \IEEEauthorblockA{
    \IEEEauthorrefmark{1}\{QiChen, SihaiTang\}@my.unt.edu,
    \IEEEauthorrefmark{2}\{Qing.Yang, Song.Fu\}@unt.edu
  }
}

\maketitle

\begin{abstract}
Autonomous vehicles may make wrong decisions due to inaccurate detection and recognition. Therefore, an intelligent vehicle can combine its own data with that of other vehicles to enhance perceptive ability, and thus improve detection accuracy and driving safety. 
However, multi-vehicle cooperative perception requires the integration of real world scenes and the traffic of raw sensor data exchange far exceeds the bandwidth of existing vehicular networks. To the best our knowledge, we are the first to conduct a study on raw-data level cooperative perception for enhancing the detection ability of self-driving systems. In this work, relying on LiDAR 3D point clouds, we fuse the sensor data collected from different positions and angles of connected vehicles. A point cloud based 3D object detection method is proposed to work on a diversity of aligned point clouds. Experimental results on KITTI and our collected dataset show that the proposed system outperforms perception by extending sensing area, improving detection accuracy and promoting augmented results. Most importantly, we demonstrate it is possible to transmit point clouds data for cooperative perception via existing vehicular network technologies.
\end{abstract}

\section{\textbf{Introduction}}

A significant part of the push towards autonomous driving vehicles, or self-driving vehicles, has been supported by the prospect that they will save lives by getting involved in fewer crashes with fewer injuries and deaths than human-driven cars. However, up until this point, most comparisons between human driven cars and self-driving vehicles have been unbalanced and contain various unfair elements.
Self-driving cars do not experience fatigue, emotional debilitation such as anger or frustration. But, they are unable to react to uncertain and ambiguous situations with the same skill or anticipation of an attentive and seasoned human driver. 

Similarly, isolated self driving vehicles may make wrong decision due to the failure of objects detection and recognition. Just as a human driver will make bad decisions while under the influence, such decisions made by the vehicle based on these failures will prove just as bad or worse than their human counterpart.
Such vehicles must completely rely on itself for decision making, and thus will not have the privilege of data redundancy, i.e., no information is received from nearby vehicles. 
Sensor failure or any other technical error will lead to fallacious results, leading to disastrous impacts. 

\subsection{\textbf{Motivations}}

The deficit of data due to single source will ultimately have a negative impact as well. Take the example of Tesla's crash in California, the car made a fatal decision because it's sensors picked up the concrete barrier but discarded the information due its immobile state on the radar
\cite{tesla_fire}. One more incident of a fatal decision is even more pronounced due to the inability to detect an vehicle from the sensors and environmental conditions. Take for example the fatal crash made by a Tesla car in Florida, where both the vehicle and the driver could not discern the white truck against a bright sky, causing the crash \cite{tesla_truck}.

Of course, there are also instance of various other circumstances leading up to bad decisions, such as the Uber training incident \cite{uber}. In this case, the vehicle did detect an unknown object, the pedestrian, from a distance. As the vehicle approached the unknown object, it gradually discerned the object to be a vehicle and finally a pedestrian, but by then, it was too late. 

We further explore the reasons why detection failure happened.
It is easy to determine that some detection failures are caused due to objects being blocked or existing in the blind zones of the sensors.
Detection failures could also be caused by bad recognition because the received signal is too weak or because the signal is missing due to system malfunction.

Our motivation comes from these incidents, because in contrast to isolated autonomous driving vehicles, like the ones in the accidents, connected autonomous vehicles (CAV) can share their collected data with each other leading to more information. We propose that information sharing can improve driving performance and experiences. Constructive data redundancy will provide endless possibilities for safe driving and multiple vehicles can collaborate together to compensate for data scarcity and provide a whole new scope for the vehicle in need. Autonomous vehicles have powerful perception systems, and together, they can achieve a proper data sharing and analysis platform to gain much more reliability and accuracy\cite{openv}.

\subsection{\textbf{Limitations of Prior Work}}
Although adding connectivity to vehicles has its benefits, it also has challenges. By adding connectivity, there can be issues with security, privacy, and data analytics and aggregation due to the large volume of information being accessed and shared. 

Current state of multi-sensor fusion consists of three distinct categories: low level fusion, feature level fusion, and high level fusion \cite{lever}. Each of these categories possess its own unique advantages and disadvantages. As their names imply, low level fusion consists of raw data fusion without any pre-processing done to the data. Feature-level fusion takes the features extracted from the raw data before fusion. Finally, high level fusion takes the objects detected from each individual sensors and conducts the fusion on the object detection results \cite{lever}.

High level fusion is often opted over the other two levels of fusion due to being less complex, but this is not suitable for our needs. Object level relies too heavily on single vehicular sensors and will only work when both vehicles share a reference object in their detection. This does not solve the issue of previously undetected objects, which will remain undetected even after fusion. And thus, we turn our sights on the other two categories.

\subsection{\textbf{Proposed Solution}}

%
To tackle the issue, we look at one of the base categories, the low level fusion of raw data.
Raw sensing data is an integral part of all sensors on autonomous driving vehicle, therefore, it is very suitable for transferring them between different cars from various manufactures. 
As such, the heterogeneity of different data processing algorithms would not affect the accuracy of the data being shared among vehicles.
As autonomous driving is of and in itself a crucial task, being so integrated in the vehicle, even a single small error in detection can lead to a catastrophic accident. Therefore, we need the autonomous cars to perceive the environment with as much clarity as possible. To achieve this end goal, they will need a robust and reliable perception system. 

Two major issues that we seek to address in doing so are as follows: (1) the type of data that we need to share among vehicles, and (2) the amount of the data that needs to be transferred versus the amount of data that is actually necessary to the recipient vehicle.
The first issue arises with the shareable data within the dataset native to the car. The second problem exists in the sheer amount of data that each vehicle generates. Since each autonomous vehicle will collect more than 1000GB of data \cite{intel} every day the challenge of assembling only the regional data becomes even harder. Similarly, reconstructing the shared data collected from different positions and angles by nearby perception system is another major challenge.

Of the different types of raw data, we propose to use the LiDAR (Light Detection and Ranging) point clouds as a solution for the following reasons:
\begin{itemize}
    \item LiDAR point clouds have the advantage of spatial dimension over 2D images and video.
    \item Native obfuscation of entities or private data such as people's faces and license plate numbers while preserving the accurate model of the perceived object.
    \item Versatility in the fusion process over images and video due to the data being consisted from points rather than pixels. For image or video fusion, the requirement is a clear zone of overlap, and this is unnecessary for point cloud data, making this a much more robust choice, especially when taking the different possible point of views of cars into perspective. 
\end{itemize}

With the three different highlights of using the raw LiDAR data as our fusion substrate, we propose the \underline{\textbf{Coo}}perative \underline{\textbf{Per}}ception (\textbf{Cooper}) system for connected autonomous vehicles based on 3D point clouds.

\subsection{\textbf{Contributions}}

Inaccurate object detection and recognition are major impediments in achieving a powerful and effective perception system. Autonomous vehicle eventually succumb to this inability and fail to deliver the expected outcome, which is unsafe to autonomous driving. 
To address these issues we have proposed a solution in which an autonomous vehicle combines its own sensing data with that of other connected vehicles to help enhance perception. We also believe that data redundancy, as mentioned, is the solution to this problem and we can achieve it through data sharing and combination between autonomous vehicles.
The proposed Cooper system can improve the detection performance and driving experience thus providing protection and safety. Specifically, we make the following contributions.

\begin{itemize}
  \item We propose the Sparse Point-cloud Object Detection (SPOD) method to detect objects in low-density point clouds data. Although SPOD is designed for low-density point cloud, it also works on high-density LiDAR data.
  \item We show how the proposed Cooper system outperforms individual perception by extending sensing area and improving detection accuracy.
  \item We demonstrate that it is possible to use existing vehicular network technology to facilitate the transmission of region of interest (ROI) LiDAR data among vehicles to realize cooperative perception.
  
\end{itemize}

\section{\textbf{Cooperative Sensing}}

Given the current outlook and work done in the field of data fusion in autonomous vehicles, we need to go a step further and define what we see as cooperative sensing. We envision cooperative sensing for CAVs as a series of challenges and benefits that will be an unavoidable part of progress.

\subsection{\textbf{Benefits of Sharing}}
Based on our observations, we wonder if detection accuracy can be improved using sensor data from multiple cars.
As we know, the sensing devices on autonomous vehicles work together to map the local environment and monitor the motion surrounding vehicles.
According to the collected data, shareable resources can be extracted from these vehicles.
For example, there is a blocked area region behind obstacles on the road that could not be sensed by one car but data gathered for this same area can be sensed and provided by other nearby cars.
Meanwhile, vehicles on adjacent districts or crowded zones can keep connection for a longer duration, thereby enhancing cooperative sensing, which will greatly help other vehicles by providing crucial information.
Hence, we propose a cooperative perception method to improve autonomous driving performance. 
This framework facilitates a vehicle to combine its sensor data with that of its cooperators' to enhance perceptive ability, and thus improving detection accuracy and driving safety. 

\subsection{\textbf{Difficulty of Sharing}}
Even though shareable resources offer useful information, vehicles prefer to utilize raw data rather than extracted results.
The detected results from other cars are hard to authenticate and trust issues further complicate this matter.
Also, since sharing all collected data is also impractical, we need to take into consideration the bandwidth and latency of vehicular networks. First, the bandwidth and latency of vehicular networks must satisfy data transmission for cooperative perception.
Then, the vehicles need to reconstruct the received data because it was taken on different positions and angles.
With this series of questions, we elaborate our research on building cooperative perception.

\subsection{\textbf{Data Choice}}
First, we demonstrate which type of sensing data is suitable for cooperative perception.
Noting that perception systems are mainly developed on image-based and LiDAR-based sensor data.
As we mention before, image data holds advantage on object classification and recognition while lacking on location information. 
In the next section, our proposed SPOD method overcomes the shortcomings of point clouds, which were too sparse to detect objects.
Based on the above reasons, we make a priority of these two sensor data for cooperative sensing. 
We prefer LiDAR data because it holds advantage in providing location information \cite{lidar}.
By only extracting positional coordinates and reflection value, point clouds can be compress into 200 $KB$ per scan.
For some applications, such as small object detection, for example license plate tracking, it is difficult for point clouds to recognize plate information.
However, when utilized with cooperative perception, we are still able to locate the plates in point clouds and ask for its image data from connected vehicles.
Because image and LiDAR point clouds are aligned together in perception system's installation, we integrate the above demand-driven strategy mainly relying on point clouds.
In some cases, it is necessary to extract a fragment of the image data in cooperative perception.

\subsection{\textbf{Data Reconstruction}}
Also, vehicles need to reconstruct the received data because it was taken on different positions and angles.
By exchanging LiDAR data, local environment can be reconstructed intuitively by merging point clouds into its physical positions.
In order to reconstruct local environment by mapping point clouds into physical positions, additional information is encapsulated into the exchange package.
Said package should be constituted from LiDAR sensor installation information and its GPS reading, which determines the center point position of every frame of point clouds.
Vehicle's IMU (inertial measurement unit) reading is also required because it records the offset information of the vehicle during driving: it represents a rotation whose yaw, pitch, and roll angles are $\alpha$, $\beta$ and $\gamma$, respectively \cite{rotation}.
A rotation matrix $R$ will be generated in Equation \ref{metric}.

\begin{equation} 
\label{metric}
R = R_{z}(\alpha )R_{y}(\beta )R_{x}(\gamma )
\end{equation}
Here $R_{z}(\alpha ), R_{y}(\beta ), R_{x}(\gamma )$ are three basic rotation matrices rotate vectors by an angle on the z-, y-, x-axis in three dimensions.

\centerline{$R_{z}(\alpha )=\begin{bmatrix} cos\alpha  & -sin\alpha  & 0\\ sin\alpha  & cos\alpha  & 0\\ 0 & 0 & 1 \end{bmatrix}$}

\centerline{$R_{y}(\beta )=\begin{bmatrix} cos\beta  & 0 &sin\beta \\0 & 1 & 0\\ -sin\beta & 0 & cos\beta \end{bmatrix}$}

\centerline{$R_{x}(\gamma )=\begin{bmatrix}  1 & 0 & 0 \\0 & cos\gamma  & -sin\gamma \\ 0 & sin\gamma  & cos\gamma \end{bmatrix}$}.

\begin{equation} 
\label{metric_union}
\begin{bmatrix} X\\Y\\ Z\end{bmatrix}=\begin{bmatrix} X_{R}\\Y_{R}\\ Z_{R}\end{bmatrix}\bigcup \begin{bmatrix} X_{T}^{'}\\Y_{T}^{'}\\ Z_{T}^{'}\end{bmatrix}
\end{equation}

\begin{equation} 
\label{metric_ratation}
 \begin{bmatrix} X_{T}^{'}\\Y_{T}^{'}\\ Z_{T}^{'}\end{bmatrix}=R\times \begin{bmatrix} X_{T}\\Y_{T}\\ Z_{T}\end{bmatrix}+\begin{bmatrix} \Delta d_{x_{T}}\\ \Delta d_{y_{T}}\\  \Delta d_{z_{T}^{}}\end{bmatrix}
\end{equation}

When connected vehicles exchange message, cooperative perception produces a new frame by combining transmitter and receiver's sensor data using Equation \ref{metric_union}, where we have the set of all coordinates equal to the coordinates of the receiver union with the the coordinates from the transmitter. However, as the transmitting vehicle is in a different state than the receiver, we must apply a transform to the original coordinates so that they match the state of the receiving vehicle.
To obtain the correct state for the transmitter's orientation, we use Equation \ref{metric}.

Note, the $X$, $Y$, and $Z$ in $\begin{bmatrix} X Y Z\end{bmatrix}{}'$ represents the 3-D space value of each point in the LiDAR point cloud data, and $\begin{bmatrix} X_{T}^{'}  Y_{T}^{'} Z_{T}^{'}\end{bmatrix}{}'$ is the transmitter's point cloud after applying the transform $R$ to the translated coordinates of the transmitting vehicle.
The transform is calculated by Equation \ref{metric}, using the IMU value difference between the transmitter and the receiver.



\section{\textbf{Cooperative Perception}}
In this section, we will show how to detect objects on cooperative sparse LiDAR point could data.

\subsection{\textbf{Object Detection based on Point Clouds}}
As we know, each self-driving vehicle will extract sensor data to perceive details in the local environment, such as lane detection, traffic sign detection and objects like cars, cyclists and pedestrians.
However, accurate detection of objects in point clouds is a challenge due to LiDAR point clouds being sparse and it having a highly variable point density.
For example, recently, based on point clouds dataset in KITTI \cite{kitti}, VoxelNet \cite{volexnet} has announced its experiments on car detection task which outperformed the state-of-the-art 3D detection methods.
Its car detection average precision is 89.60\%, and for smaller objects, such as pedestrians and cyclists, the average precision drops to 65.95\% and 74.41\% respectively in a fully visible (easy) detecting environment.
While in a difficult to see (hard) detecting condition, the car, pedestrian and cyclist detection further drop to 78.57\%, 56.98\%, and 50.49\%, respectively.
Another insight here is that LiDAR provides sparse 3D point clouds with location information but is hard to classify and recognize. 
To analyze the results from the above works, we cannot ignore the failure detection.
This allows us to approach the issue from another perspective - cooperative sensing methods to improve detection accuracy.

\subsection{\textbf{Sparse Point-cloud Object Detection (SPOD)}}
Typically autonomous vehicles use single end-to-end deep neutral network to operate on a raw point cloud.
However, after cooperative sensing, the re-constructed data from different LiDAR devices may have different features like point density.  
For example, Velodyne \cite{wiki-velo} produces 64-beam, 32-beam and 16-beam LiDAR devices, which provide different density point clouds.
Similar to image's resolution, 3D detector using deep neutral network may have inaccuracy recognition results when used on low density point clouds.
We note that 64-beam LiDAR, which provide the highest resolution LiDAR data, is well adopted by researches and companies on 3D object detection \cite{volexnet,second}. 
While some others, as in our case, use 16-beam LiDAR, which outputs sparse data but has a price advantage over its higher end counterparts. 
This requires our proposed detection method on its assembled 3D detection model not only to work on high density data, but also can detect objects from much sparser point clouds. 
Unfortunately, these convolutional neural network (CNN)-based object detection methods are not suitable for low-density data because of insufficient of input features.
Inspired by ~\cite{second} proposed SECOND, an end-to-end deep neural network that learns points-wise features from point clouds, we propose the Sparse Point-cloud Object Detection (SPOD) methods which can adapt low density point clouds.

\subsection{\textbf{Architecture of SPOD}}
\begin{figure}[h]
\centering
\includegraphics[width=0.38\textheight]{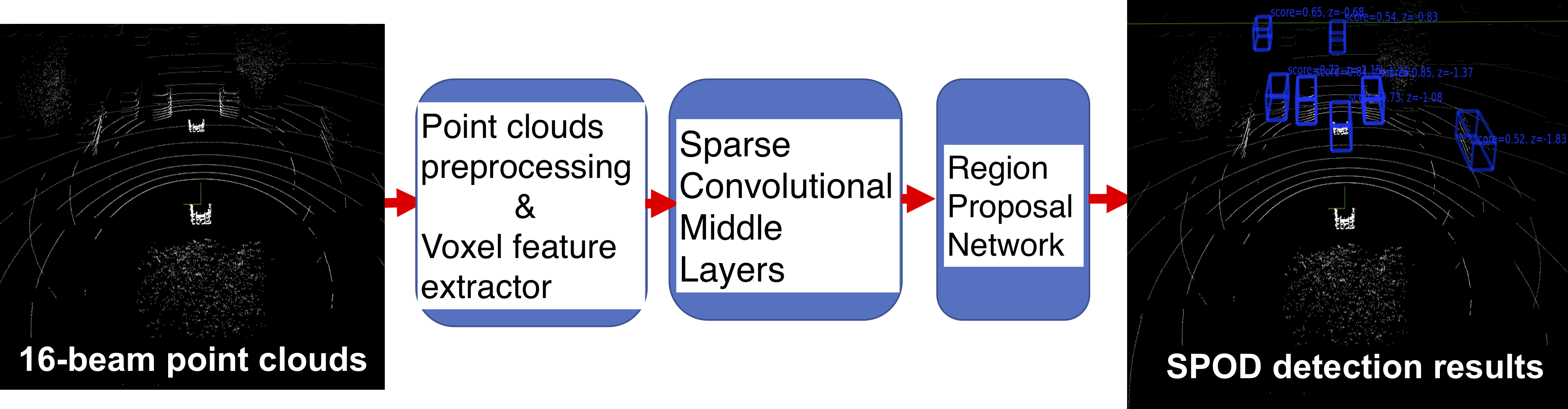}
\caption{Structure of the SPOD 3D object detection method.}
\label{figure:detector}
\end{figure}

The proposed detector, depicted in Fig. \ref{figure:detector}, consists of three components.
Our adopted 3D LiDAR point cloud is represented as a set of cartesian coordinates, (x, y, z) with reflection values.
The distribution of point clouds is much too sparse and irregular.
Specifically in the preprocessing, to obtain a more compact representation, point clouds are projected onto a sphere using approach from \cite{squeeze} to generate a dense representation.
In voxel feature extractor components, our framework takes represented point clouds as input, feeding extract voxel-wise features to voxel feature encoding layer, this is well demonstrated by Voxelnet \cite{volexnet}.
Then a sparse convolutional middle layer \cite{scnn} is applied. 
Sparse CNN offers computational benefits in LiDAR-based detection because the grouping step for point clouds will generate a large number of sparse voxels. 
In this approach, output points are not computed if there is no related input points.
Finally, Region Proposal Network (RPN) \cite{rpn} is constructed using single shot multibox detector (SSD) architecture \cite{ssd}. 
The feature maps as input to RPN from Sparse CNN and are concatenate into one feature map for prediction.
Framework in every vehicle use this single end-to-end trainable network to produce 3D detection results not only from dense LiDAR data but also from low resolution LiDAR data from nearby vehicles.
\FloatBarrier
\FloatBarrier 
Eventually, we successfully adopt SPOD to detect objects both on our collected sparse data and on dense KITTI data.
In the next section, we demonstrate a full evaluation of SPOD detection.

\section{\textbf{Evaluation and Result Analysis}}
In this section, we evaluate the performance of the proposed Cooper system using two real-world LiDAR datasets.

\subsection{\textbf{Datasets}}

In the experiment, we test Cooper on two datasets: the KITTI dataset provided by the Karlsruhe Institute of Technology and Toyota Technological Institute at Chicago, and T$\&$J dataset collected by our semi-autonomous driving golf cart.
Therefore, we obtain two types (dense and sparse) of point clouds.
In the dense KITTI dateset, a 64-beam LiDAR sensor is used to collect point clouds.
But in our T$\&$J dataset, which supplies 16-beam point cloud, the collected point cloud is 4X more sparse than KITTI's, of course, the amount of data is 4X decreased respectively.
With the two datasets, we then fully evaluate the performance of the Cooper system for a total of 19 scenarios.
Based on the KITTI testset data, we choose four different sets of road driving test scenarios. 
And at the same time, in order to enrich the experimental content and verify our design effects, we conduct 15 experiments on Cooper using the T$\&$J dataset.
\textit{Note that Cooper can also be applied to heterogeneous point clouds input.}
We elected not to conduct this test due to a lack of suitable LiDAR datasets.

We define \textit{single shot} as point clouds collected by an individual vehicle, and \textit{cooperative sensing} as merging all point clouds from nearby vehicles.
We systematically analyze the test results of single shot and cooperative sensing to demonstrate the performance improvement on object detection.
Qualitative results of Cooper under two experimental datasets are demonstrated in the following sections.

\subsection{\textbf{Evaluations on KITTI Dataset}}
\begin{figure*}[h]
\centering
      \begin{subfigure}[b]{0.28\textwidth}
                \includegraphics[width=\textwidth]{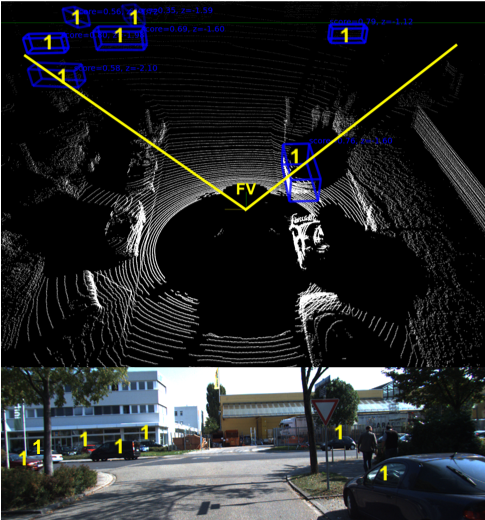}
                \caption{Single shot at $t1$: a vehicle utilizes SPOD on 64-beam point clouds to detect cars, and the results are shown in blue boxes.}
                \label{figure:64-t1}
                \end{subfigure}
        \begin{subfigure}[b]{0.28\textwidth}
                \includegraphics[width=\textwidth]{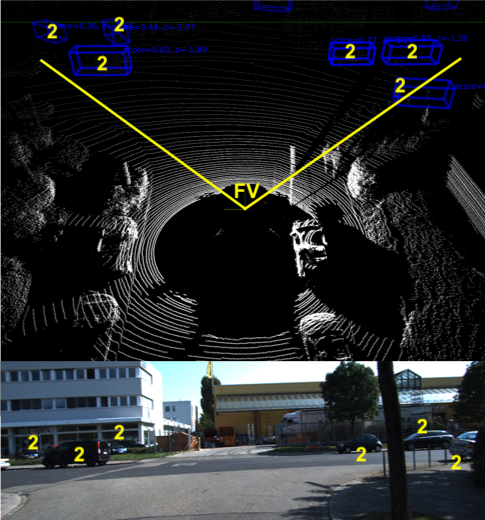}
                \caption{Single shot at $t2$: as the vehicle moving forward, its detection results are drawn in blue boxes. Bottom image provides the ground truth.}
                \label{figure:64-t2}
                \end{subfigure}
        \begin{subfigure}[b]{0.35\textwidth}
                \includegraphics[width=\textwidth]{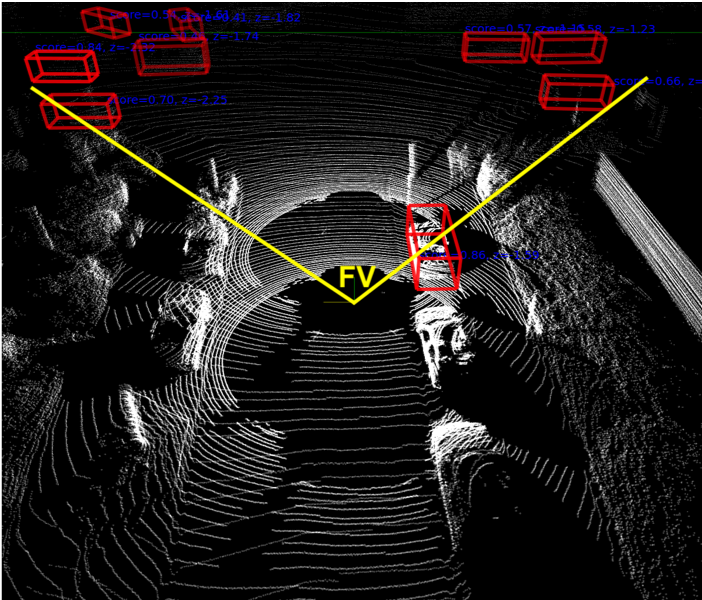}
                \caption{Merging $t1$ and $t2$’s point clouds to produce cooperative point clouds. The detected cars are drawn in red boxes using the same SPOD detector.}
                \label{figure:64-com}
                \end{subfigure}
        \caption{Cooperative detection of vehicles based on the KITTI point clouds.}
        \vspace{-11pt}
        \label{fig:64}
\end{figure*}

In this section, we evaluate Cooper's performance using the KITTI dataset.
As we know, KITTI provides raw consecutive 3D Velodyne point clouds in several scenarios.
We choose one such segment of sensing data in folder $2011/09/26/0009$ as an example, shown in Fig. \ref{fig:64}.
\FloatBarrier
\FloatBarrier 
To corresponding with $120 \degree$ front view image, this LiDAR data of front-view area is evaluated. 
At beginning time $t1$, one single shot frame of 64-beam raw point cloud is collected in Fig. \ref{figure:64-t1}.
As the testing vehicle is moving forward after two seconds, another single shot frame of 64-beam raw point cloud is collected at time $t2$ shown in Fig. \ref{figure:64-t2}.
By merging $t1$ and $t2$'s point clouds, we emulate the cooperative sensing process between two vehicles. 
We utilize SPOD object detector to detect cars and draw results in red boxes to bound detected cars in Fig. \ref{figure:64-com}
Meanwhile, in order to compare the detection results on Cooper, we also apply SPOD on single shot point clouds collected at times $t1$ and $t2$. 
The detected cars are drawn in blue boxes, as shown in Fig. \ref{figure:64-t1} and Fig. \ref{figure:64-t2}, repectively.
From the figures, we can observe two major improvements of employing cooperative perception. 
\textbf{First, the sensing range is extended by data sharing.}
We can see that at $t1$ we observe 6 blue boxes, and at $t2$ we observe 6 blue boxes yet again. However, when combined, we observe a total of 9 detected cars (red boxes) in the merged data, which include all the cars detected at $t1$ and $t2$.
\textbf{Second, the detecting score/confidence value of some detected vehicles is increased. }
For example, a vehicle in Fig. \ref{figure:64-t1} is detected with a detecting score of $0.76$ at $t1$, and the same vehicle is also detected in Fig.~\ref{figure:64-com}, but the detecting score of this vehicle is increased (by 13\%) to $0.86$.
We also provide the corresponding images as the ground truth at the bottom of Fig. \ref{figure:64-t1} and Fig.~\ref{figure:64-t2}.

\begin{figure}[h]
\centering
\includegraphics[width=0.38\textheight]{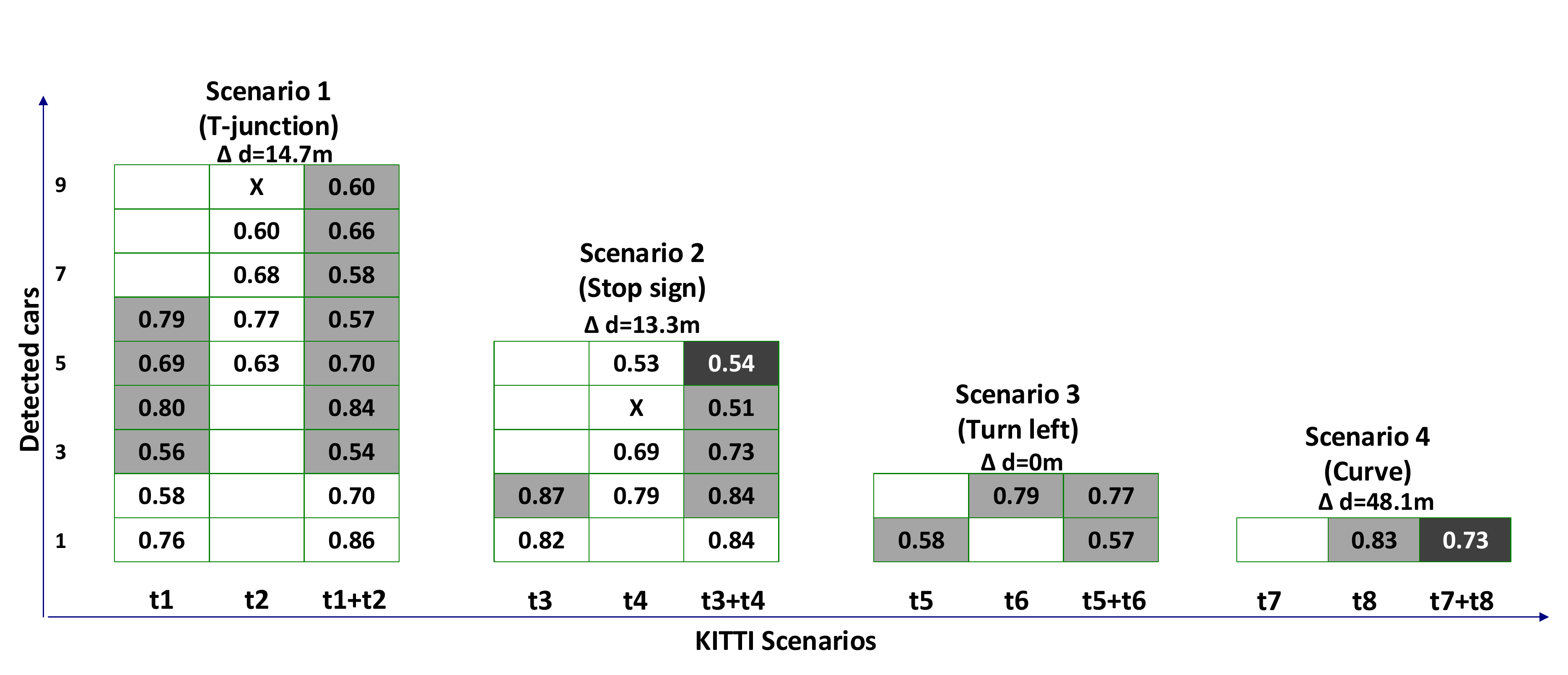}
\caption{Vehicle detection results in four different scenarios in KITTI.}
\label{figure:kitti_dis}
\end{figure}

\begin{figure}[h]
\centering
\includegraphics[width=0.3\textheight]{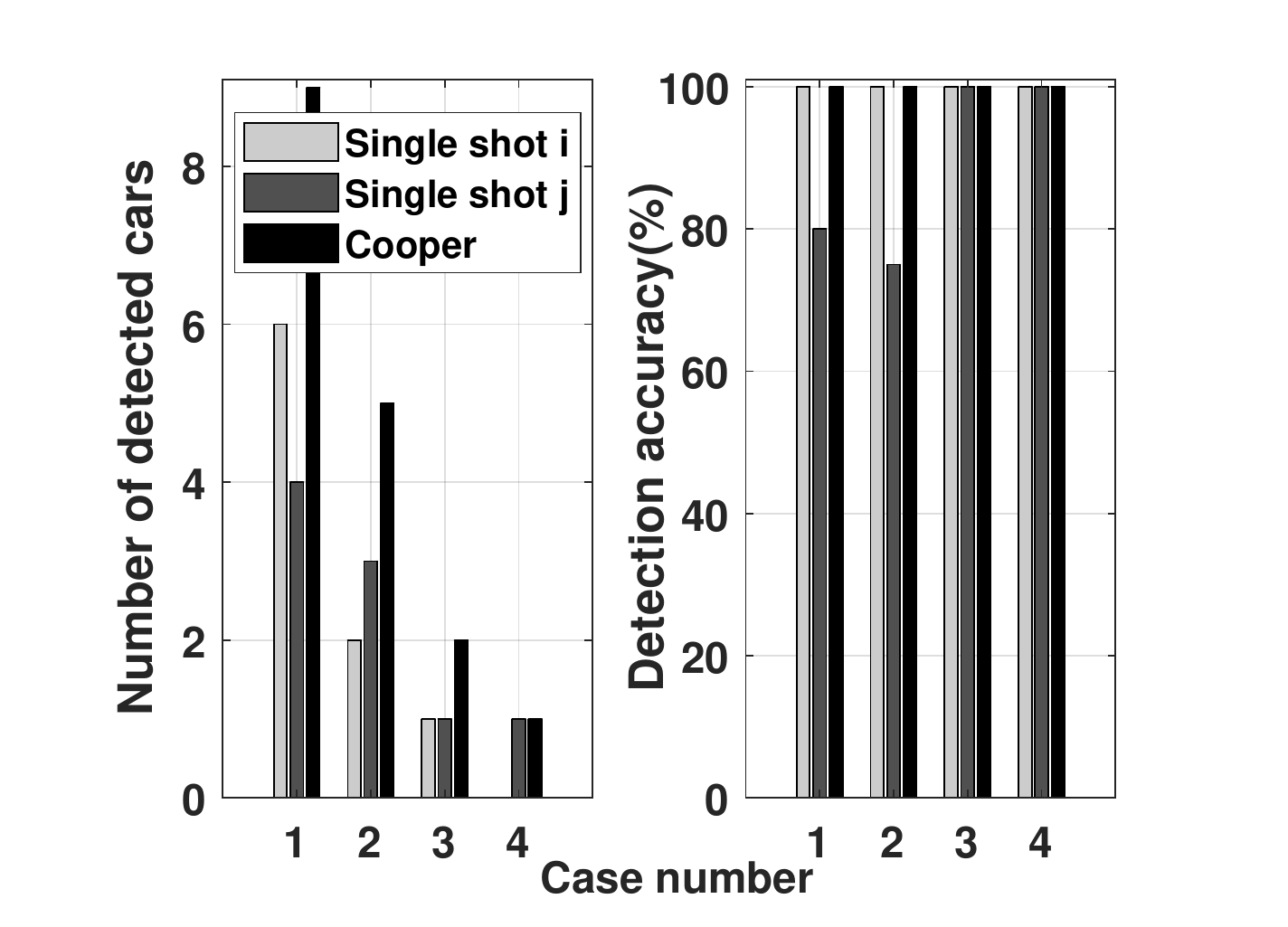}
\caption{Number of cars detected and the detection scores in four KITTI scenarios.}
\label{figure:ki_det}
\end{figure}

The following is calculating the number of vehicles detected by single shot and cooperative sensing in four different scenarios: T-junction, stop sign, left turn and curve scenarios. 
The single shot data collected by two vehicles are labeled as $t1$ and $t2$, $t3$ and $t4$, $t5$ and $t6$, $t7$ and $t8$ in four scenarios, respectively.
Therefore, the data marked as $t1 + t2$, $t3 + t4$, $t5 + t6$, and $t7 + t8$ are the cooperative data, combining the single shot point clouds.
We then compare the vehicle detection results against the ground truth (captured in images) for each case, and depict the results in Fig. \ref{figure:kitti_dis}.
The value of $\Delta d$ indicates the distance between the two locations of the vehicle at two different times.
Every three columns represents a cooperative process, which is similar to the example we demonstrated in Fig. \ref{fig:64}.
We draw the distribution of detection results using cells in each column. 
The number in each cell is the detecting score, the higher the score, the more positive the result.
The symbol $X$ represents a missing detection, i.e., the detecting score is too low.
The cell without score means the object is out of detection area.
Also, different colors are used to indicate the distance.
The darker the color, the farther the distance. 
According to the actual detection distance of LiDAR, we divide it into three scales of near (<10m), medium (10-25m) and far (>25m), which are represented in the illustration by white, gray and black, respectively.
It is clear that the amount of detected cars in cooperative data is equal to or exceeds the number in individual single shots. 
Then, we use qualitative results to analyze the performance on the number and accuracy of detected vehicles shown in Fig. \ref{figure:ki_det}.
\textbf{The proposed Cooper method not only detects more cars, but also grants better detection accuracy because there is no missing detection in the cooperative point clouds.}

\subsection{\textbf{T$\&$J Dataset}}
Unfortunately, KITTI dataset does not provide enough experimental scenarios because it is a vision benchmark collected by isolated instances of single vehicles. 
We are committed to multi-vehicle cooperation, and thus, to improve the driving safety and experience of CAV, we build a dataset that is suitable for vehicle collaboration, naming it the T$\&$J dataset.

Our testing cars are equipped with high precision sensing systems, such as LiDAR system, radar system, vision system, and supplemental system such as GPS and IMU sensors. More specifically our sensor framework consists of the following sensors:
\begin{itemize}
  \item  {2 X Front-view cameras}
  \item  {4 X Surround-view fish eye cameras}
  \item{1 X Inertial and GPS sensor}
  \item  {1 X Front-view 120$\degree$ Radar}
  \item  {1 X Velodyne VLP-16 360$\degree$ LiDAR}
  \item  {1 X Nvidia PX2}
\end{itemize}

Velodyne VLP-16 360 LiDAR \cite{wiki-velo} is used along with Radar, which utilizes radio waves to measure distance. LiDAR provides low resolution image information.
Cameras, on the other hand, provides very high resolution image information, but, it fails to perform in extreme weather or environmental conditions. Four fish-eye lens cameras are used to perceive and navigate the surrounding environment.
IMU sensors provides the system that monitors the dynamically changing movements of the vehicle. 
Also, GPS sensor data can be used to obtain a rough estimate of the location or the positioning of the car.  
Nvidia Drive PX2 \cite{px2} is a scalable AI supercomputer for our autonomous driving. 

\subsection{\textbf{Evaluation on T$\&$J Dataset}}
\begin{figure*}[h]
\centering
      \begin{subfigure}[b]{0.25\textwidth}
                \includegraphics[width=\textwidth]{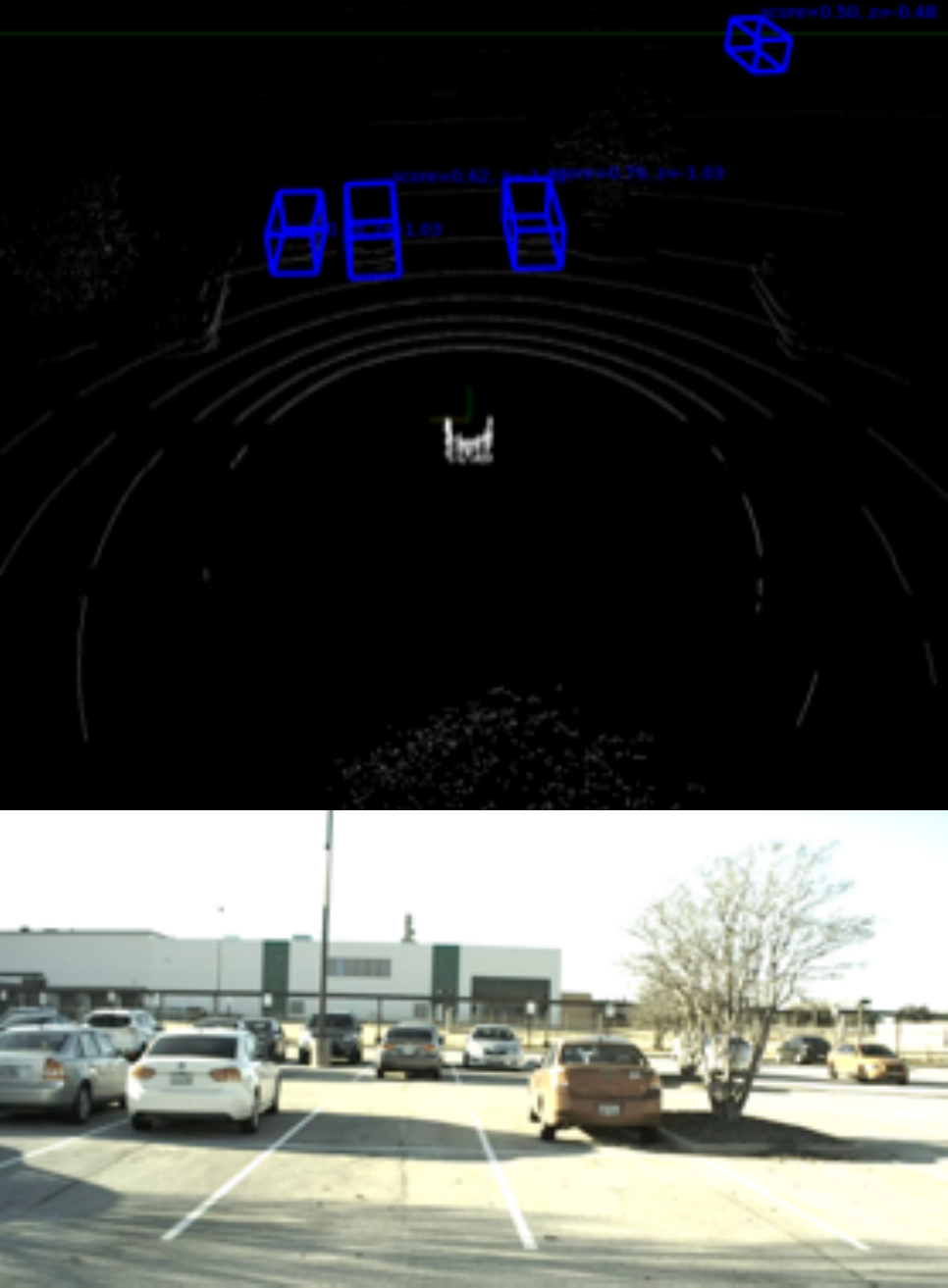}
                \caption{Single shot at $t1$: applying SPOD on 16-beam point clouds to detect cars.}
                \label{figure:16-t1}
                \end{subfigure}
        \begin{subfigure}[b]{0.25\textwidth}
                \includegraphics[width=\textwidth]{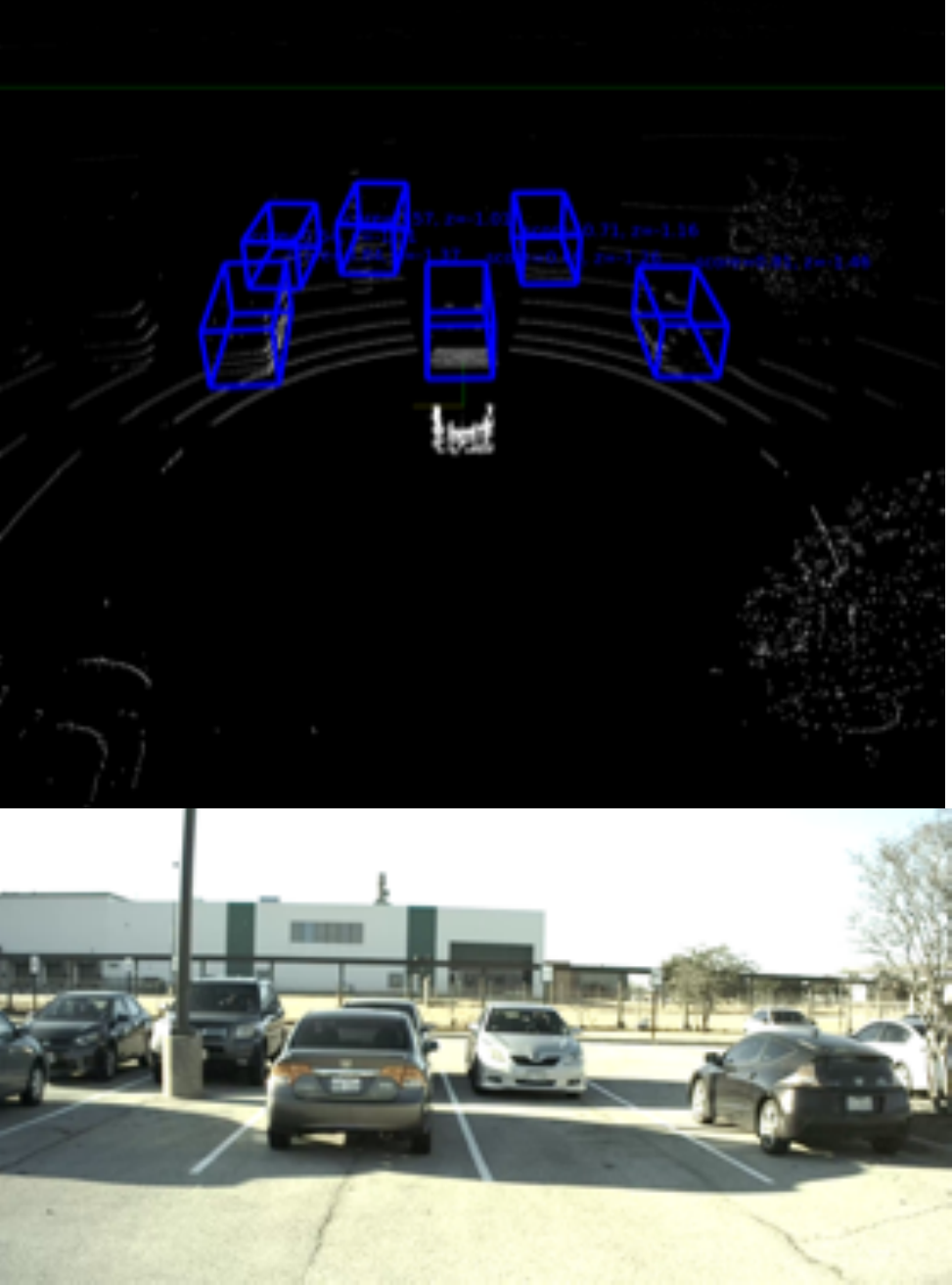}
                \caption{Single shot at $t2$, vehicle detection results are shown in blue boxes. }
                \label{figure:16-t2}
                \end{subfigure}
        \begin{subfigure}[b]{0.35\textwidth}
                \includegraphics[width=\textwidth]{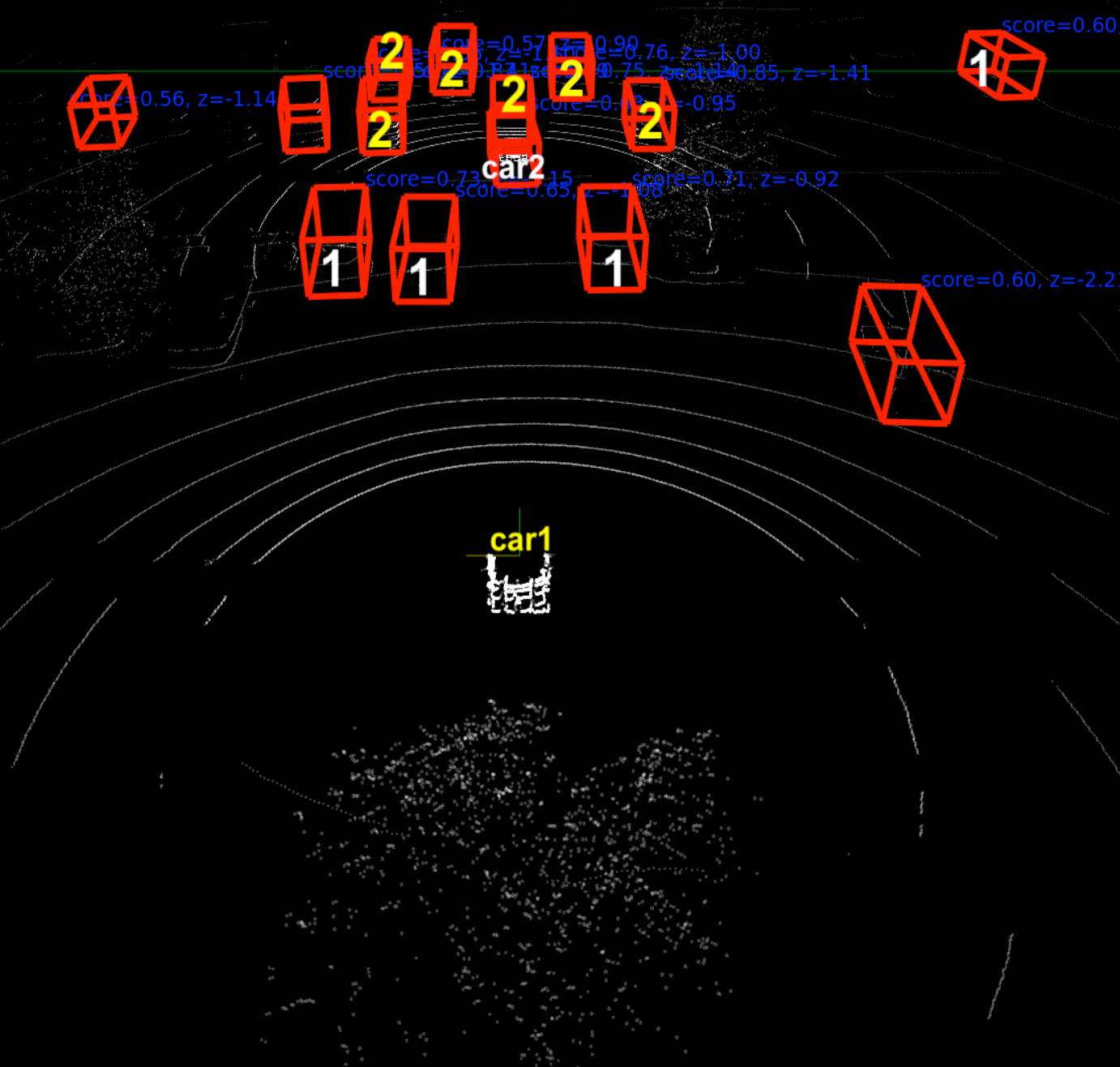}
                \caption{Cooperative perception combines two single shots. The detected cars are drawn in red boxes using the SPOD object detector.}
                \label{figure:16-com}
                \end{subfigure}
        \caption{An example of cooperative perception using the T$\&$J dataset.}
        \label{fig:16}
\end{figure*}

We evaluate Cooper's performance on our T$\&$J dataset.
We select a sequence of continuous frames of front-view LiDAR point clouds and show them in Fig. \ref{fig:16}.
It can be clearly found that our point cloud is much more sparse than that from KITTI.
All the data in the T$\&$J dataset are collected in a parking lot.
A frame of 16-beam raw point cloud data is shown in Fig. \ref{figure:16-t1}.
Another collected single shot data is shown in Fig. \ref{figure:16-t2}.
By merging these two frames of point clouds, we produce two vehicles’ cooperative sensing. 
3D detector detects cars and draws results in red boxes to bound detected cars' location in Fig. \ref{figure:16-com}.
Similar to Fig. \ref{fig:64}, SPOD detects cars in two single shots and draws them in blue boxes. SPOD also draws results in red boxes to bound detected cars in cooperative sensing. 
Meanwhile, ground truth images are shown at the bottom of Fig. \ref{figure:16-t1} and Fig. \ref{figure:16-t2}.
By studying this case, we conclude that sensing area is expanded by data sharing because Fig. \ref{figure:16-com} detects all the objects in the single shots.
Most importantly, we see that the presence of new cars are discovered, cars that were not presence in the previous single shots.
\textbf{This phenomenon is a direct proof to the shortcomings of fusion on object level. Due to neither vehicles detecting the objects by themselves, there stands no possible way for the object-level fusion to detect the objects that were missed. This, we avoid and overcome with low level fusion.}

We marked the cars detected at time $t1$ and $t2$ by numbers 1 and 2, respectively. 
It is worth noting that there are three unmarked vehicles in Fig. \ref{figure:16-com}.
This is a significant discovery as this phenomenon indicates an \textit{increase in the detection capability of cooperative perception}.
We can extrapolate and assume that by receiving the perceptual information from nearby vehicles, Cooper can greatly enhance a vehicle's range of perception, allowing for better detection of traffic information.

\begin{figure*}[h]
\centering
      \begin{subfigure}[b]{0.45\textwidth}
                \includegraphics[width=\textwidth]{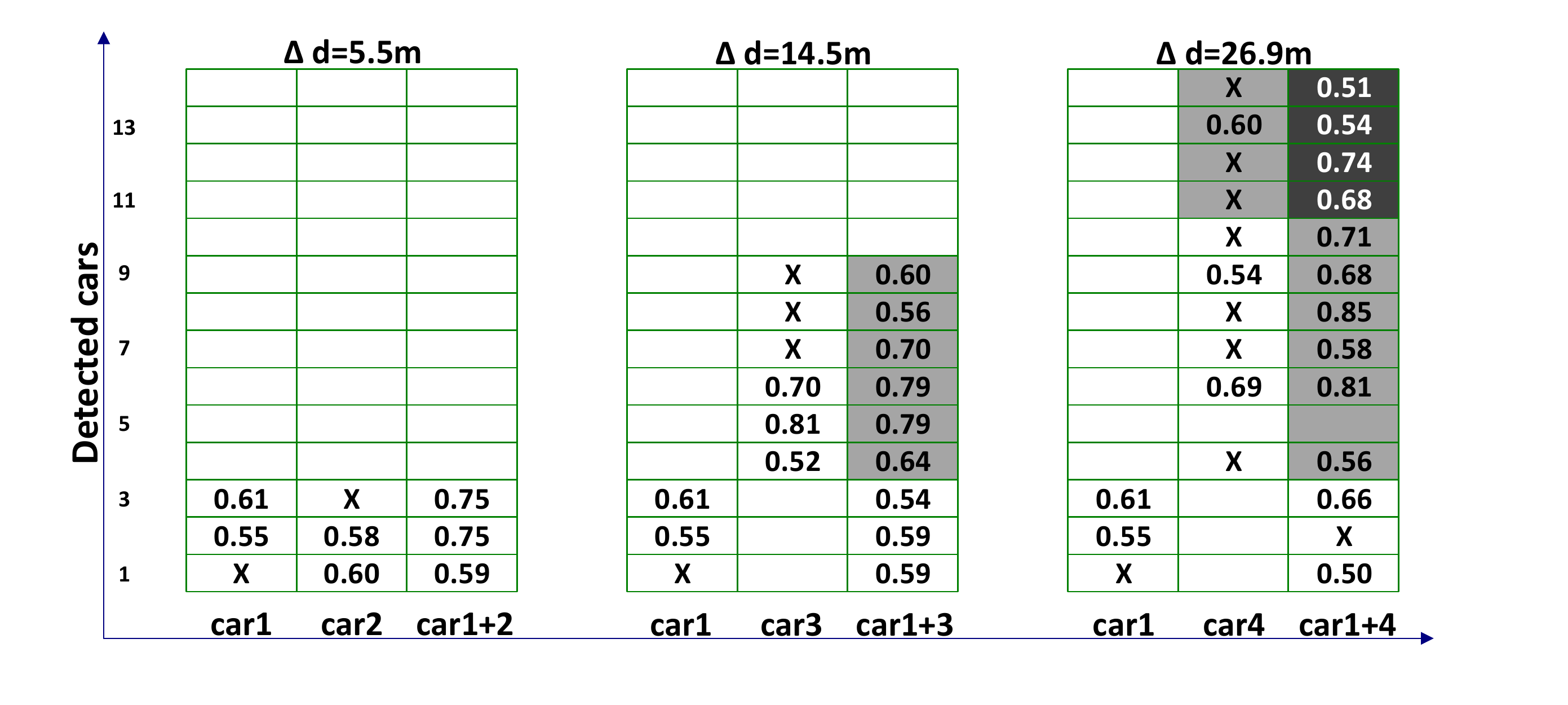}
                \caption{Scenario 1}
                \label{figure:tj1_dis}
                \end{subfigure}
        \begin{subfigure}[b]{0.45\textwidth}
                \includegraphics[width=\textwidth]{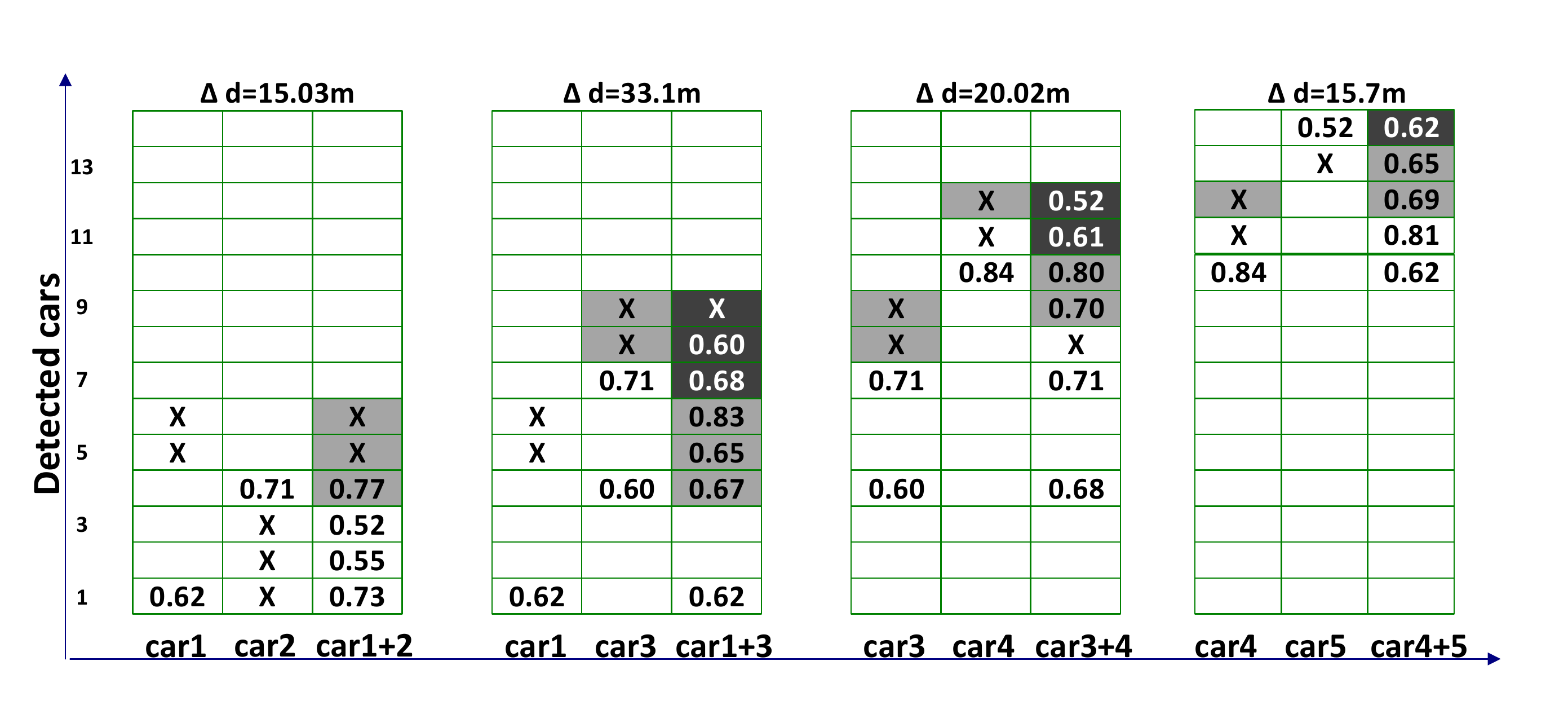}
                \caption{Scenario 2}
                \label{figure:tj2_dis}
                \end{subfigure}
        \begin{subfigure}[b]{0.45\textwidth}
                \includegraphics[width=\textwidth]{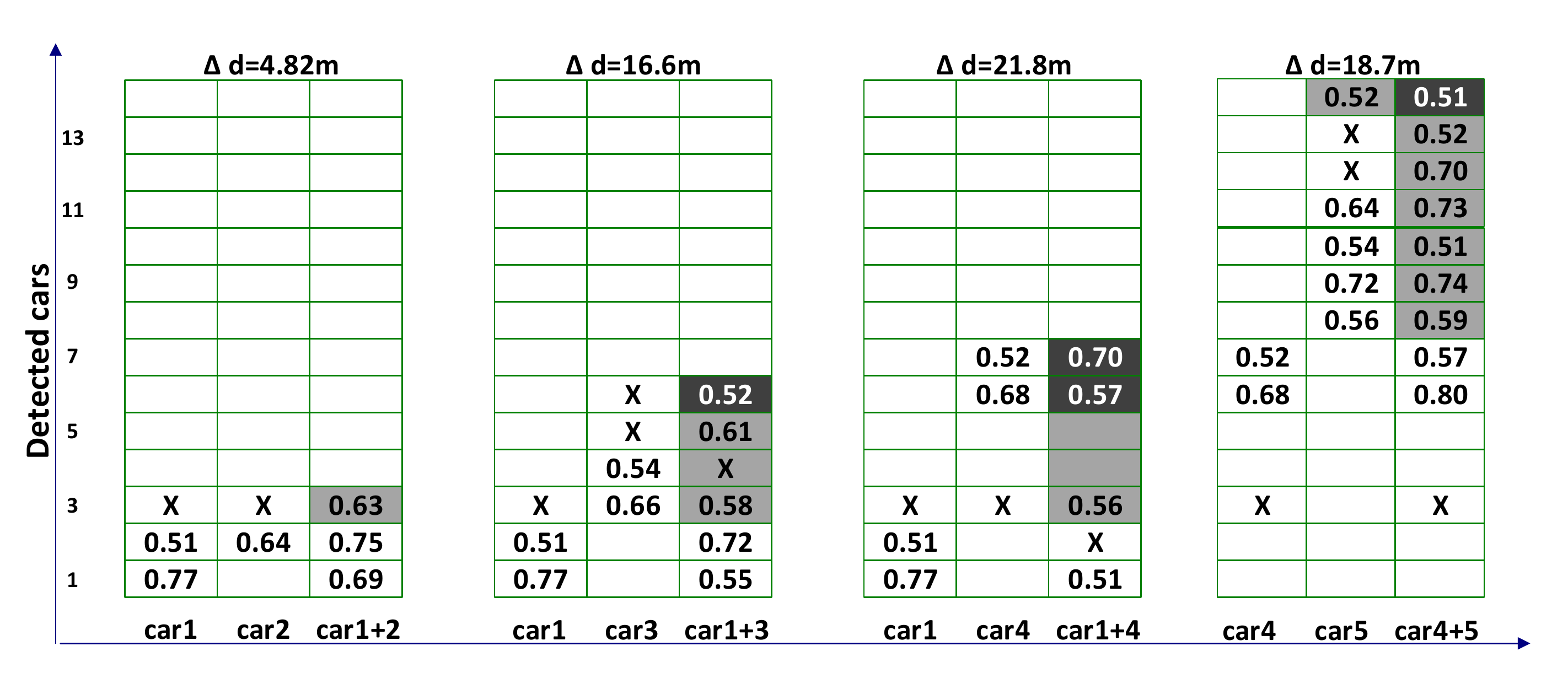}
                \caption{Scenario 3}
                \label{figure:tj3_dis}
                \end{subfigure}
        \begin{subfigure}[b]{0.45\textwidth}
                \includegraphics[width=\textwidth]{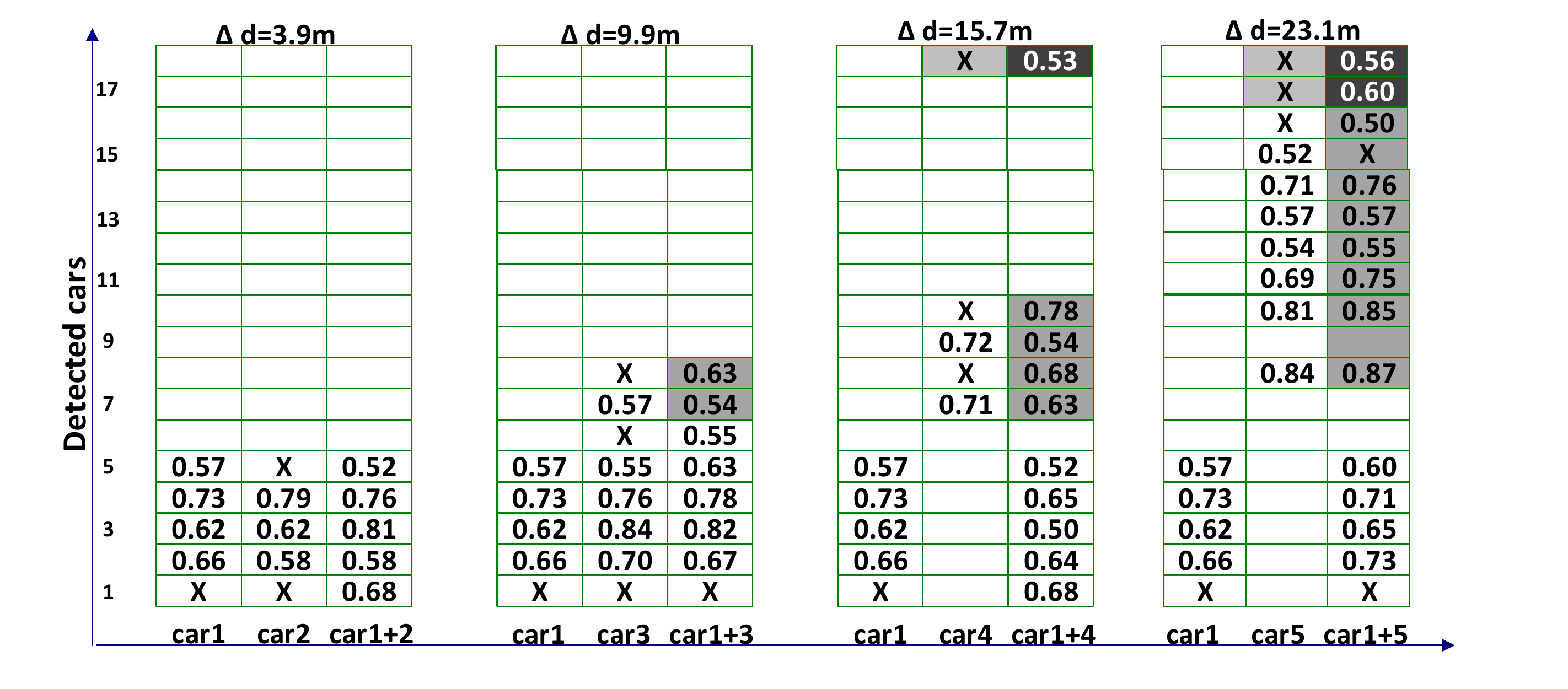}
                \caption{Scenario 4}
                \label{figure:tj4_dis}
                \end{subfigure}
\caption{Vehicle detection results in different scenarios in the T$\&$J dataset.}
\label{figure:tj_dis}
\end{figure*}

The T$\&$J dataset provided four sets of testing data, which were collected on the roads around our campus's parking lots.
In the four scenarios, we conduct cooperative perception experiments. 
Different from the KITTI, in each experimental scenario, we sample the fusion data at different distances, so as to better display the disparity of information collected by vehicles in different regions.
As Fig. \ref{figure:tj_dis} shows, in each scenario, we list detailed detection results of Cooper at different distances.
Similar to Fig. \ref{figure:kitti_dis}, every three columns corresponding the SPOD detection results on two single shots and one cooperative sensing, represents a cooperative perception case.
The test car can receive both nearby sensing data and relatively long-distance sensing data. 
For example, in Fig. \ref{figure:tj1_dis}, from left to right, there are three cases in which a vehicle cooperates with other three located at three distances.
It can be seen that in the cooperative perception of adjacent areas, such as the left cases in Scenarios 1 and 4, the individual detection results of two single shots are similar, but both output undetected targets, because these targets are blocked by unknown means in the single shots.
Through cooperative perception, point clouds of blocked area are supplemented by each other, thereby these targets are detected.
Moreover, the detected targets in both cases, after cooperative perception, have a marked increased in detecting scores.
We evidence this phenomenon due to the redundancy of data and the presence of more features is gathered by harvesting detailed point clouds. 

In all scenarios shown in Fig. \ref{figure:tj_dis}, we carry out the cooperative perception of two cars, both are relatively far apart from each other.
As a result, the detection area is expanded even lager.
\FloatBarrier 
\FloatBarrier
\noindent Every car can detect the target in front of itself.
But for distant targets, they are powerless due to scarcity or blockage of point clouds.
Cooperative perception enables global detection of objects located at far, medium, and near distance.
Objects are appeared in cells of different colors.
Similarly, some objects that are undetected by single shot are detected in cooperative sensing.
This reinforces the fact that some objects that were not detected through traditional means can be discovered through data fusion.
This shows that our design can complement some key features.
This is a significant discovery on cooperative perception.

\begin{figure}[h]
\centering
      \begin{subfigure}[b]{0.24\textwidth}
                \includegraphics[width=\textwidth]{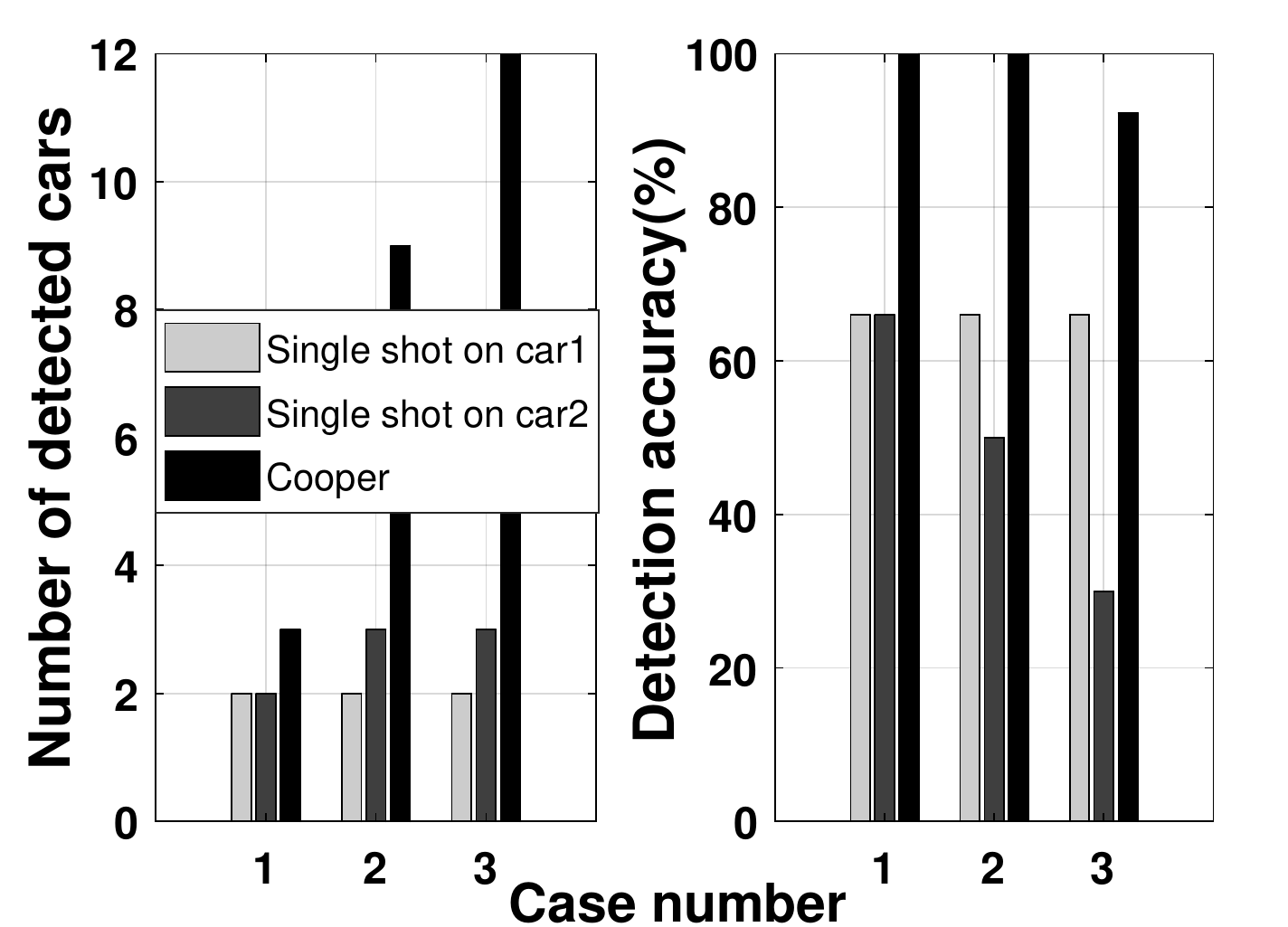}
                \caption{Scenario 1}
                \label{figure:tj1_det}
                \end{subfigure}
        \begin{subfigure}[b]{0.24\textwidth}
                \includegraphics[width=\textwidth]{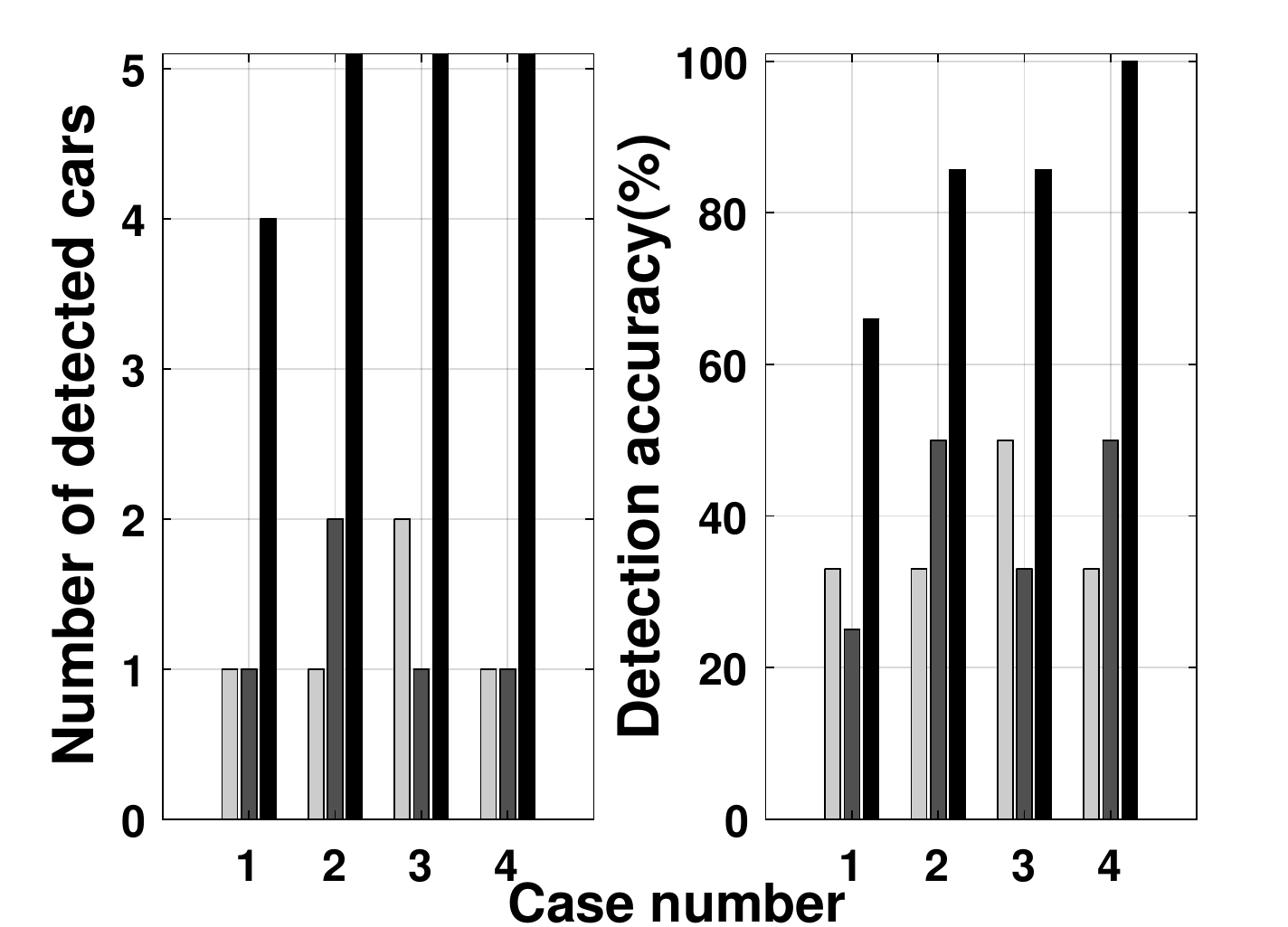}
                \caption{Scenario 2}
                \label{figure:tj2_det}
                \end{subfigure}
        \begin{subfigure}[b]{0.24\textwidth}
                \includegraphics[width=\textwidth]{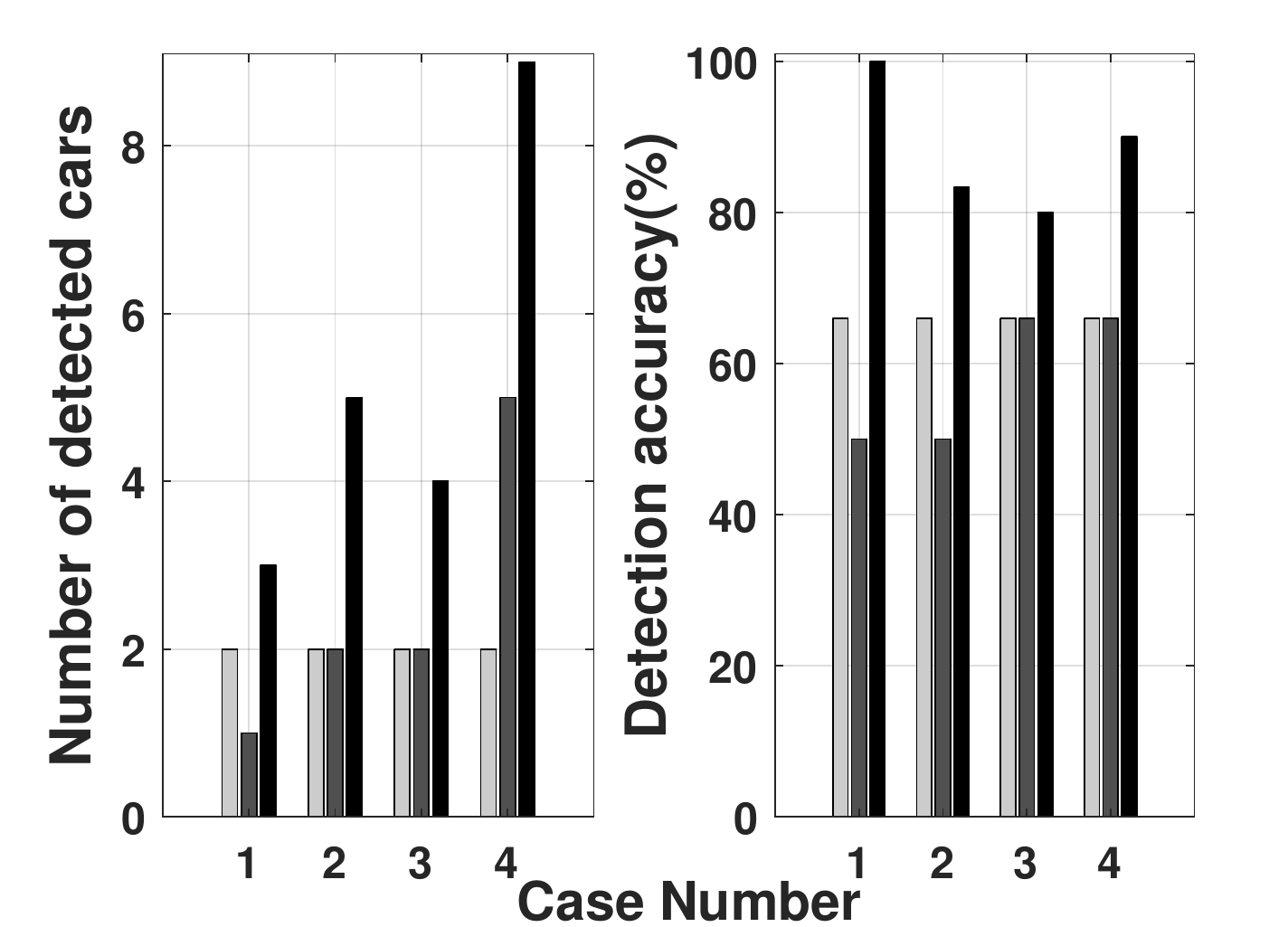}
                \caption{Scenario 3}
                \label{figure:tj3_det}
                \end{subfigure}
        \begin{subfigure}[b]{0.24\textwidth}
                \includegraphics[width=\textwidth]{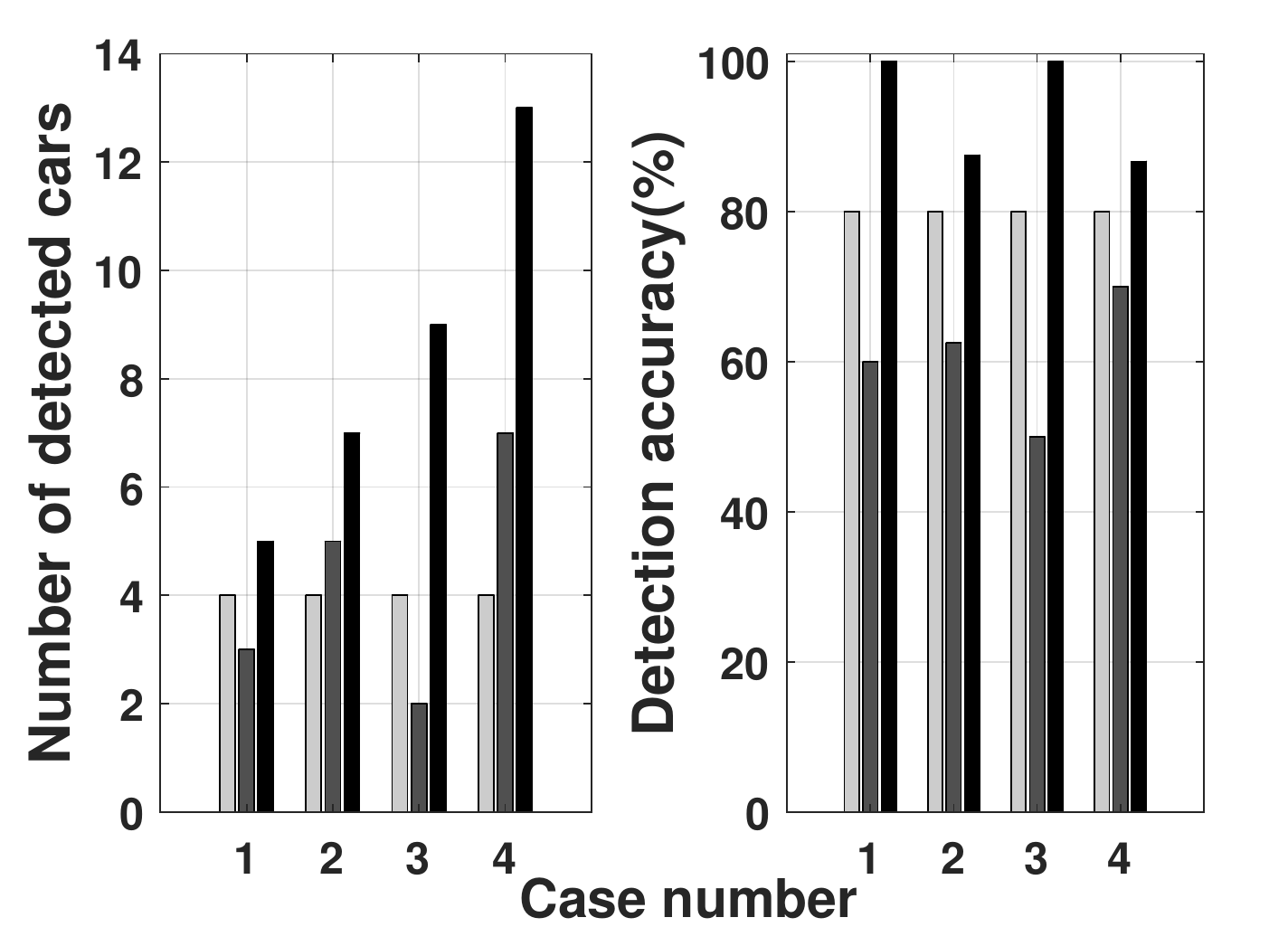}
                \caption{Scenario 4}
                \label{figure:tj4_det}
                \end{subfigure}
\caption{Details on the detection results on the T$\&$J dataset.}
\label{figure:tj_det}
\end{figure}

Then, we use qualitative results to analyze the performance on the number and detecting scores of detected vehicles, shown in Fig. \ref{figure:tj_det}.
From Scenario 1, we have the single shot analysis results for three different cases. It is clear that the number of cars detected based on the fused data is much higher than either of the cars alone. Despite the high detection rate, we do see that even while fused there are still some cars not being detected. 

In Scenario 2, we find that there is a high amount of cars that is hard to detect from either car alone, but shows up when fused. This change of environment hold high relevance to common place areas such as a full parking lot or congested junctions where each car is limited by the cars around it. Should there be a speeding car that is ignoring stop signs or running the red light, the fusion will mitigate the likelihood of a missed detection for all cars involved in the immediate vicinity. 

In both Scenarios 3 and 4, we find that, similar to the trend shown in Scenarios 1 and 2, we have a closely related relationship between fusion and increase in object detection. As each scenario takes place in different environments, time of the day, different levels of congestion, the fusion method is proven robust and is able to adapt to different environments while retaining its capabilities to augment the status quo.

\subsection{\textbf{Statistical Analysis}}
Our statistical analysis results show that in the experimental scenarios of KITTI and T$\&$J datasets, some of the targets in cooperative perception are detected by both, some by only one, and some are detected by neither.
Detection difficulty is thereby classified as easy, moderate and hard, respectively. Specifically, easy refers to when one or more vehicles are able to detect the same object. Moderate refers to when only one vehicle is able to clearly detect this object. Finally, hard is given when no cars are able to detect this object.

\begin{figure}[h]
\centering
\includegraphics[width=0.33\textheight]{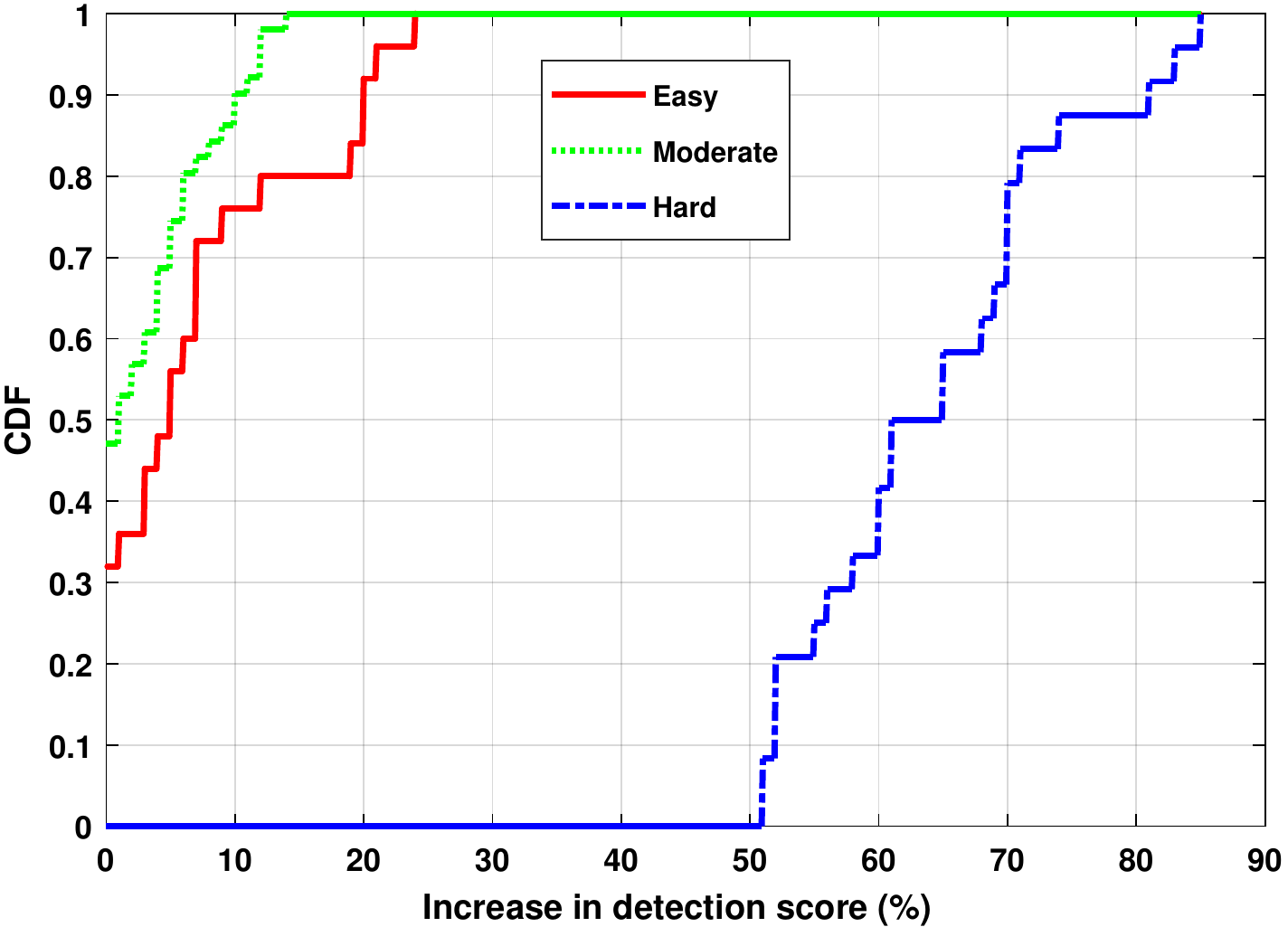}
\caption{Improvement of the detection performance by cooperative perception.}
\label{figure:cdf}
\end{figure}

In Fig. \ref{figure:cdf}, we calculate the improvement of detection performance on these three types of objects. For example, from the line marked easy, we see that we have an improvement of 10\% in detection score for 80\% of the time. Taking the direct implication of our testing, we see that the detection scores for easy and moderate achieve a marginal yet consistent increase in detection rate; mainly distributed within 10\% in detection score improvement.
This is because both easy and moderate objects contain detailed and saturated sized point clouds captured from a single scene, resulting in the fusion providing only marginal improvements to the detection results.

However, note that when we test the third type of objects, the hard objects detected by neither, we find that we are consistent with our findings that we have above, \textbf{our detection score improvement is a flat increase of 50\% in raw detection score at worst and just this alone is enough for autonomous vehicles to note the object for avoidance prevention, because they only need to know that there is an object there where previously one was not discovered.}

\begin{figure}[h]
\centering
\includegraphics[width=0.3\textheight]{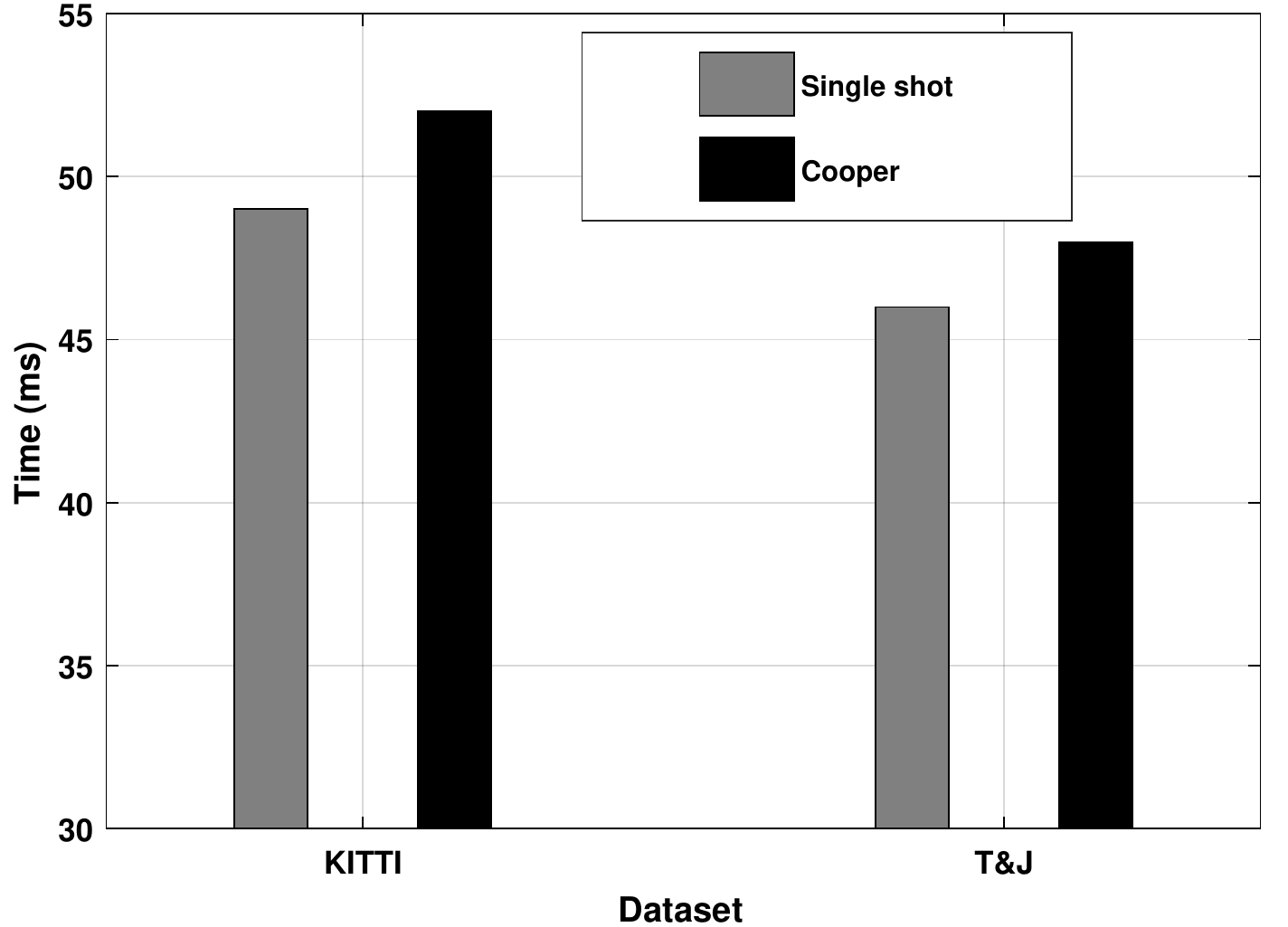}
\caption{Time needed to detect objects on single shot and cooperative sensing data.}
\label{figure:time}
\end{figure}

We record time cost of detection based on single shot and cooperative data, shown in Fig. \ref{figure:time}.

As latency impacts the performance and reliability of all autonomous vehicles heavily, we tested our fusion method against both the KITTI data and our own. In the test, the SPOD model for 3D car detection is executed on a computer with a GeForce GTX 1080 Ti GPU \cite{wiki-Nvidia}.
In both experiments, we compared the time needed for object detection in both single shots against the fused data. \textbf{In both instances, fusing the data used 5 ms over the baseline data, a very minimal increase in detection time for a significant increase in the number and accuracy of objects detected.}

\subsection{\textbf{Fusion Robustness}}
From a realistic standpoint, we will inevitably have to deal with sensor drift, so to deal with this phenomenon, we must test our fusion method of robustness against sensor drift.
When integrating GPS and IMU, we observe yields of less than 10 cm in positional errors \cite{GPS}. To test the robustness of our fusion method, we conducted procedural artificial skewing of our GPS readings. We skew the GPS data as follows:
\begin{itemize}
    \item Skewing both $x$ and $y$ coordinates to the maximum bounds of known GPS drifting.
    \item Skewing just one axis to the limit of GPS drifting.
    \item Pushing past that boundary by doubling the maximum GPS drifting to simulate abnormal instances.
\end{itemize}

\begin{figure}[h]
\centering
\includegraphics[width=0.33\textheight]{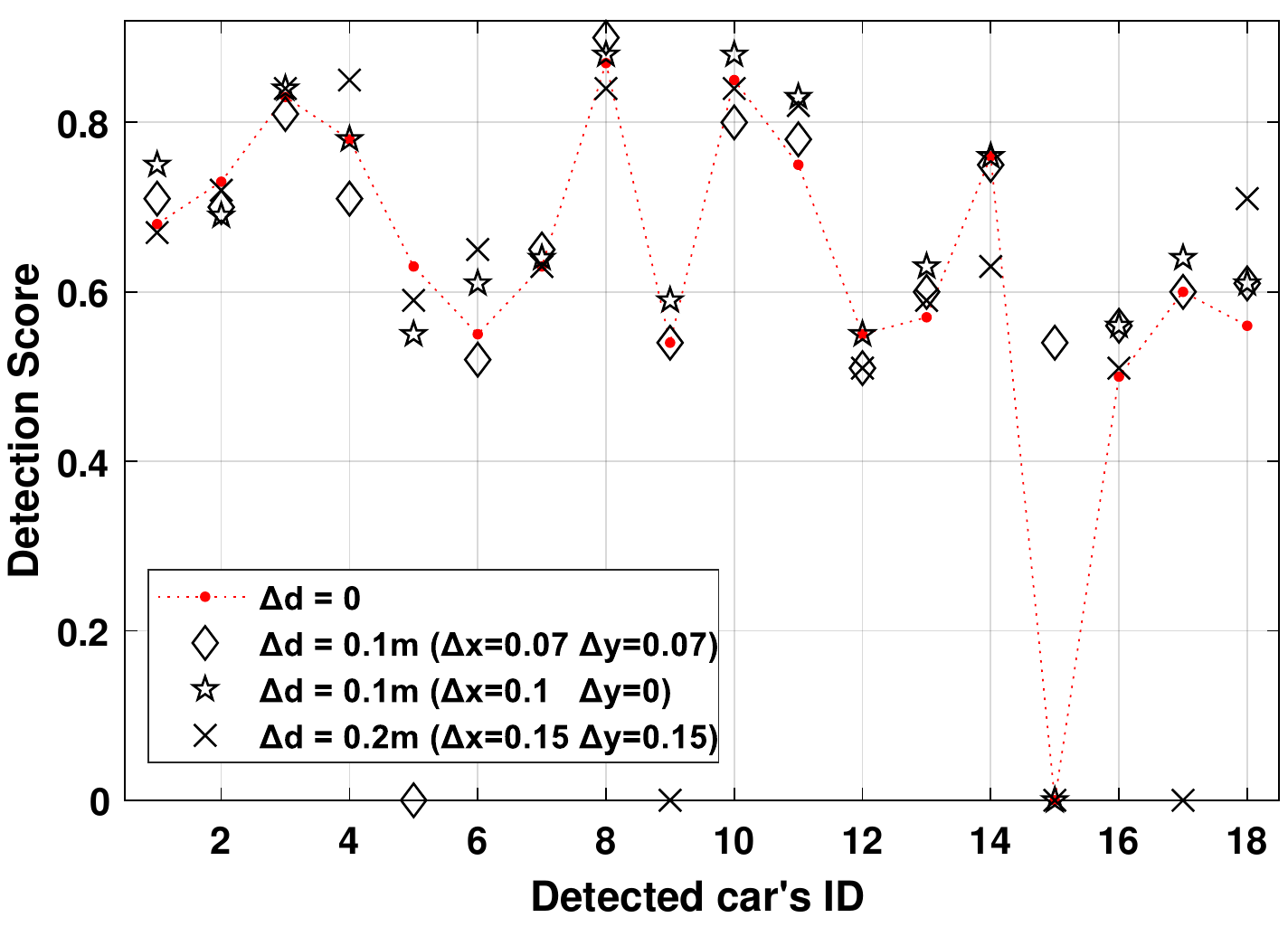}
\caption{Cooperative perception results on GPS reading drifting.}
\label{figure:GPS_det}
\end{figure}

With the GPS readings skewed, we then tested the detection score for each of the different type of drifting scenarios against the baseline GPS reading. As evinced from Fig.\ref{figure:GPS_det}, we see that with the exception of already known undetected vehicles, we have a similar clustering of the skewed detection scores versus the baseline score, with the overwhelming majority achieving successful detection. 
\textbf{It should be noted, however, that skewing the readings surprisingly improved the detection score in several instances, possibly masking the inherent drift from the baseline GPS reading. And just as some of the skewing helped the result, it also caused the detection to fail for two instances. }

\subsection{\textbf{Networking Requirements}}
Even though point clouds can be simplified to coordinate values, we still need to consider the gap between data generated by autonomous vehicles and the limited wireless networking throughput, such as the limited bandwidth provided by DSRC \cite{dsrc}.
We adopt a strategy to extract data based on the region of interest (ROI), e.g., traffic lights, blocked areas, nearby vehicles and free-space in driving path, to further reduce data size to hundreds $KB$ per frame.
Background data like buildings, trees are subtract because these information can be constructed by each vehicle after several times mapping measurement. This allows for retention of valuable information of immobile objects while keeping the size of the ROI data small. 
For object detection purpose, ROI data will be extracted whenever failure detection happened on this area. 

\begin{figure}[h]
\centering
\includegraphics[width=0.27\textheight]{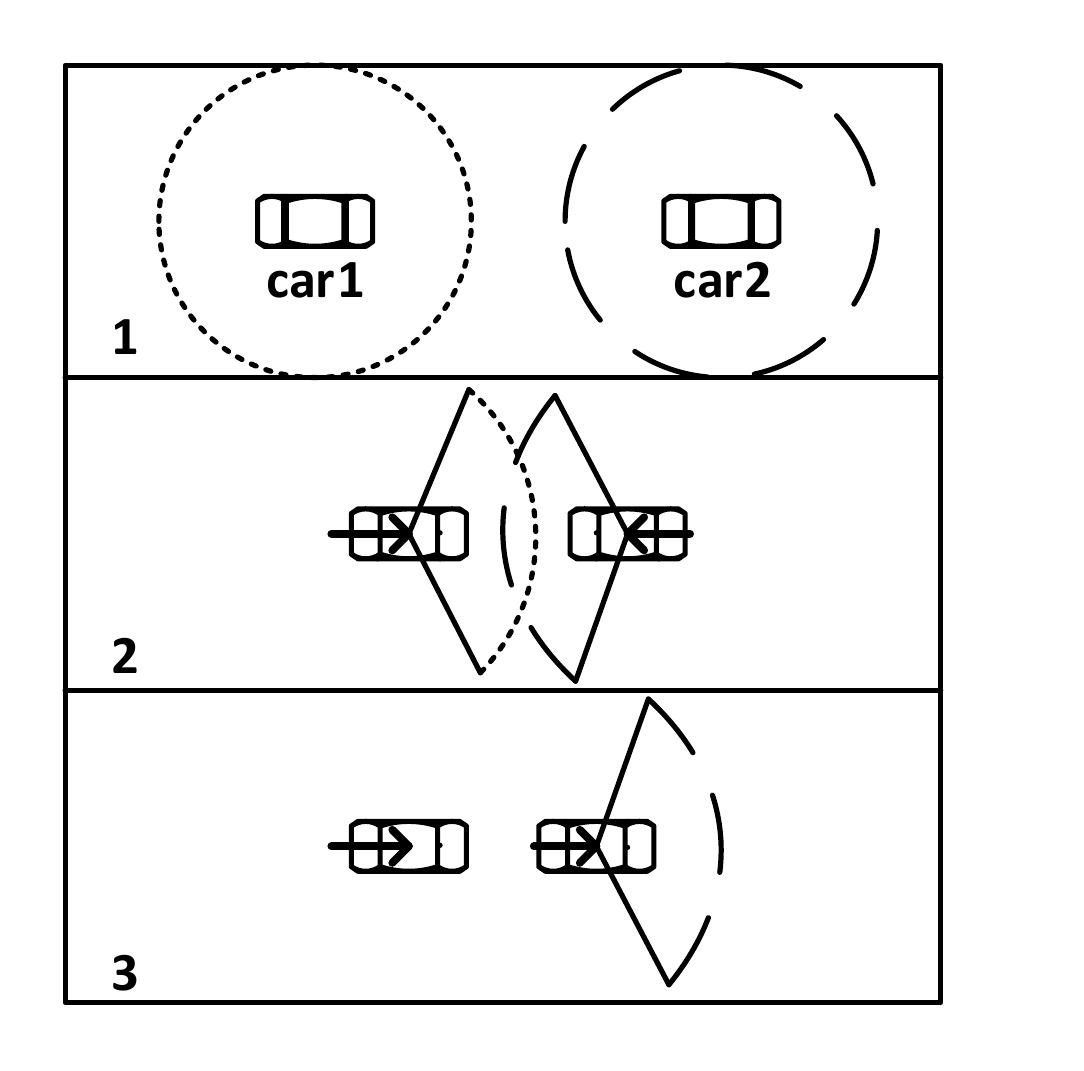}
\caption{Illustration of ROI data exchange between two vehicles.}
\label{figure:car}
\end{figure}

\begin{figure}[h]
\centering
\includegraphics[width=0.27\textheight]{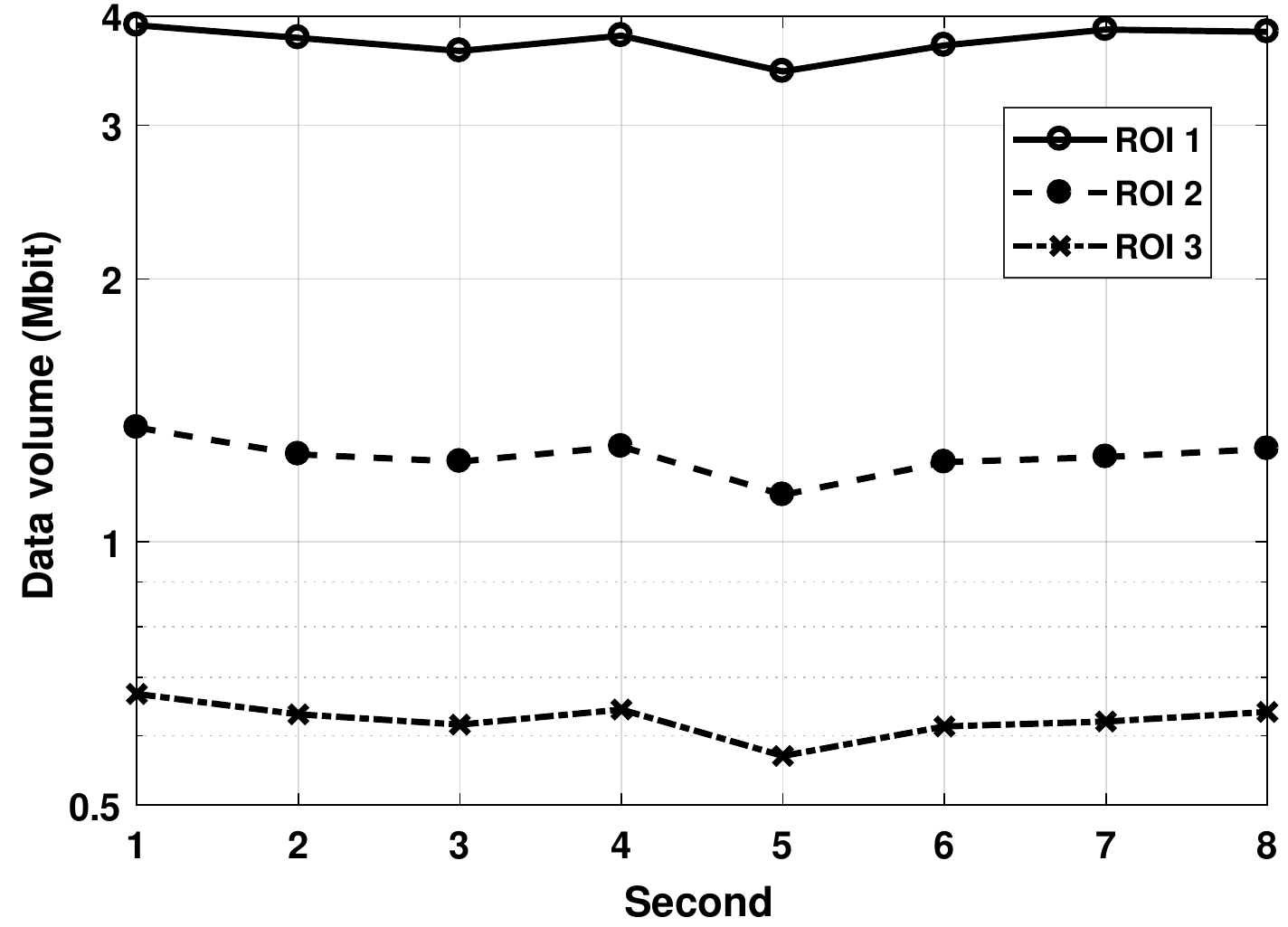}
\caption{Volume of LiDAR data being exchanged between two cars.}
\label{figure:volume}
\end{figure}

However, just knowing the relative ROI is not optimized enough. The ideal case is to have a multitude of real world ROI categories that provide a guideline for the bases of how much data is needed for an optimal balance of data size versus detection accuracy. To illustrate the importance of this tradeoff, we present three different types of ROI categories and their respective data consumption via Fig.\ref{figure:car} and Fig.\ref{figure:volume} respectively where the sample rate in the latter is 1Hz, or 1 frame per second.
We simulated and gathered the total data consumption between two cars, both utilizing a 16-beam LiDAR, every second over an eight second time frame.
Note, we observe that message exchange rate for cooperative perception does not require as high a sensing rate as the standard rate for individual vehicles.
Because for easy or moderate objects, detailed sized point clouds are already captured.
While due to blocking or distance, we may experience an insufficiency of point clouds, making objects hard to detect. 
In most cases, the native data on a recipient vehicle only needs to be supplemented by a single data frame from different view perspective.
Excessive exchanging of frequencies only leads to unnecessary data, hence needlessly congesting the communication channels. 
With efficiency and lightweight traffic as a constraint, we decided that a sample rate of 1 frame per second is enough to satisfy the needs of Cooper whilst remaining within our set of constraints.

As seen in Fig.\ref{figure:car}, we have three different scenarios, each representing a general phenomenon. For the first one, we see that two cars are fairly apart from each other, laterally but fairly close horizontally. We would typically see this situation in two lane drive with opposite directions separated by a single lane divider. In this scenario, we would ideally want as much information as possible as we lack the safety of a physical buffer between the vehicles. In situations like this, we transfer the entirety of the frame of LiDAR data and this is the most costly of all scenarios as evinced by Fig.\ref{figure:volume}. From the same scenario, we can calculate that for the most expensive data transaction, the total data size can be compress into around 1.8 $Mbit$ per frame for each car.

Next, we have the case of cars in closer lateral proximity to each other, representing typical junctions where cars from all directions are able to see the opposing car. In situations such as this, the ROI is typically the field of view from the driver's perspective, making only a 120 degree field of view our minimal requirement. As both vehicles needs to exchange this information, the transaction cost is additive for each of the participating vehicles.

Lastly, we have the most common situation of one car needing the field of view of a leading car. The trailing car is the one needing the information and thus the transaction is one way, consuming the least amount of bandwidth out of all three scenarios. 

Thus, deriving from the simulation of the three different cases, the three presented are within the capacity of DSCR bandwidth, as seen in real-world test \cite{moveset}.

\subsection{\textbf{Summary of Experiment Results}}
In summary, we prove that Copper method outperforms individual perception on extending sensing area, improving detection accuracy and complementing object detection.
We find that collaboration offers more information, even some are not perceived by individuals.
The most important, we conduct a feasibility study and demonstrate that the bandwidth of DSRC could satisfy point clouds transmission for cooperative perception.
We would like to mention that our design succeeds in privacy preservation because only LiDAR data is involved for sharing.

\section{\textbf{Related Work}}
Rapid development of autonomous vehicles has motivated research institutions to develop representations to perceive local environment, such as lane detection, traffic sign detection and detect objects like cars, cyclists and pedestrians \cite{yolo, rrc, second, volexnet} based on the open datasets \cite{city, kitti, apollo}. 
As we know, these datasets are collect by multiple sensing devices from individual vehicles.   
To achieving self-driving, we put heavy emphasis on accuracy cognition of the surrounding local environment.
However, the detection results still has vast room for improvement even when utilizing state-of-the-art Convolutional Neural networks (CNNs) \cite{cnn}.

Current works make use of low level fusion of sensors to extract the features or objects for purpose of tracking \cite{lowfuse}. However, this does not incorporate the use of raw data as is for the purpose of fusion and object detection. Papers such as \cite{midfuse} and \cite{highfuse} discuss methods of fusion that constructs theoretical architecture for low level fusion and detection. Other approaches, such as \cite{pointfusion} and \cite{f-pointnet}, proposed 3D object detection methods by fusing both image and point cloud from the same vehicle. Differing from improving the detection methods for a single vehicle, we focus our method on the fusion of data between different vehicles.

To the best of our knowledge, there are no prior work done to implement the concept of multi vehicular raw sensor data for the purpose of object detection.
This room for improvement is also the cause of severe consequences because self-driving cars may make wrong decisions due to failure of detection of objects.
A Cooper framework for connected autonomous vehicles can solve the aforementioned issues through cooperative sensing.
However, none of the public datasets and related detection methods explicitly consider low level fusion approach as a solution.

\section{\textbf{Conclusions}}
We propose Cooper for connected autonomous vehicles as an entry to a broader platform for CAV.
This method facilitates a CAV capable vehicle to combine its sensing data with that of its cooperators to enhance perceptive ability, thereby improving detection accuracy and driving safety.
In order to reconstruct local environment, we map point clouds into their corresponding object positions. This will merge and align the shared data that is collected from nearby vehicles, which may provide data scopes coming from different positions and angles.
We incorporate deep learning based SPOD with Cooper to detect 3D objects from aligned LiDAR data, marking and discovering previously undetected objects. 
Finally, we evaluated Cooper on KITTI and our collected dataset, showing that the Cooper is capable of enhancing detection performance by expanding the effective sensing area, capturing critical information in multiple scenarios and improving detection accuracy.

%
\bibliographystyle{abbrv}


\bibliography{sigproc}  
%
%

\end{document}